\documentclass[onecolumn]{doublecol-new}
\usepackage[linesnumbered]{algorithm2e}
\usepackage{makecell}
\usepackage{array}
\usepackage{stfloats}
\usepackage{subfigure}
\usepackage[subfigure]{graphfig}
\usepackage{multirow}
\usepackage{graphicx}
\usepackage{setspace}
\usepackage{pbox}
\usepackage{acro}

\usepackage[backend=bibtex,style=authoryear]{biblatex}
\bibliography{IEEEfull,lib.bib}

\DeclareAcronym{ml}{short= ML,long=Machine learning,
	class= abbrev}
\DeclareAcronym{svms}{short=SVMs,long=Support Vector Machines,
	class= abbrev}
\DeclareAcronym{pca}{short=PCA,long=Principal Component Analysis,
	class= abbrev}
\DeclareAcronym{pls}{short=PLS,long=Partial Least Squares,
	class= abbrev}
\DeclareAcronym{fpm}{short=FPM,long=First Principal Model,
	class= abbrev} 

\DeclareAcronym{ann}{short=ANN,long=Artificial Neural Network,
	class= abbrev}
\DeclareAcronym{fnn}{short=FNN,long=Feedforward Neural Network,
	class= abbrev}
\DeclareAcronym{rbfn}{short=RBFN,long=Radial Basis Function Network,
	class= abbrev}
\DeclareAcronym{rnn}{short=RNN,long=Recurrent Neural Network,
	class= abbrev}
\DeclareAcronym{fm}{short=FMs,long=Fuzzy Models,
	class= abbrev}
\DeclareAcronym{ts}{short=T-S,long=Takagi-Sugeno,
	class= abbrev}
\DeclareAcronym{fnn-1}{short=FNN,long=Fuzzy Neural Network,
	class= abbrev}

\DeclareAcronym{pso}{short=PSO,long=Particle Swarm Optimization,
	class= abbrev}
\DeclareAcronym{ga}{short=GA,long=Genetic Algorithm,
	class= abbrev}
\DeclareAcronym{cg}{short=CG,long=Conjugate Gradient,
	class= abbrev}
\DeclareAcronym{ll}{short=LL,long=Log-Likelihood,
	class= abbrev}
\DeclareAcronym{nll}{short=NLL,long=Negative value of Log-Likelihood,
	class= abbrev}  
\DeclareAcronym{map}{short=MAP,long=Maximizing A Posterior,
	class= abbrev}
\DeclareAcronym{mc}{short=MC,long=Monte Carlo,
	class= abbrev}
\DeclareAcronym{mcmc}{short=MCMC,long=Markov Chain Monte Carlo,
	class= abbrev}
\DeclareAcronym{sor}{short=SoR,long=Subset of Regressors,
	class= abbrev}
\DeclareAcronym{ivm}{short=IVM,long=Informative Vector Machine,
	class= abbrev}

\DeclareAcronym{gp}{short=GP,long=Gaussian Process,
	class= abbrev}
\DeclareAcronym{dgp}{short=DGP,long=Dependent Gaussian Process,
	class= abbrev}
\DeclareAcronym{cgp}{short=CGP,long=Convolved Gaussian Process,
	class= abbrev}
\DeclareAcronym{igp}{short=IGP,long=Independent Gaussian Process,
	class= abbrev}
\DeclareAcronym{gprn}{short=GPRN,long=Gaussian Process Regression Networks,
	class= abbrev}
\DeclareAcronym{lvm}{short=GP-LVM,long=Gaussian Latent Variable Model,
	class= abbrev}
\DeclareAcronym{lgp}{short=LGP,long=Local Gaussian Process,
	class= abbrev}
\DeclareAcronym{gpdm}{short=GPDM,long=Gaussian Process Dynamical Model,
	class= abbrev}

\DeclareAcronym{mimo}{short=MIMO,long=Multiple-Input Multiple-Output,
	class= abbrev}
\DeclareAcronym{miso}{short=MISO,long=Multiple-Input Single-Output,
	class= abbrev}
\DeclareAcronym{siso}{short=SISO,long=Single-Input Single-Output,
	class= abbrev}
\DeclareAcronym{nlti}{short=NLTI,long=Nonlinear Time-Invariant,
	class= abbrev}
\DeclareAcronym{nltv}{short=NLTV,long=Nonlinear Time-Varying,
	class= abbrev}
\DeclareAcronym{lti}{short=LTI,long=Linear Time-Invariant,
	class= abbrev}
\DeclareAcronym{ltv}{short=LTV,long=Linear Time-Varying,
	class= abbrev}

\DeclareAcronym{cep}{short=CEP,long=Certainty Equivalence Principle,
	class= abbrev}
\DeclareAcronym{uav}{short=UAV,long=Unmanned Aerial Vehicle,
	class= abbrev}
\DeclareAcronym{gmv}{short=GMV,long=Generalized Minimum Variance,
	class= abbrev}

\DeclareAcronym{cp}{short=CP,long=Convolution Process}
\DeclareAcronym{lmc}{short=LMC,long=Linear Model of Coregionalization,
	class= abbrev}
\DeclareAcronym{idc}{short=IDC,long=Inverse Dynamics Control,
	class= abbrev}
\DeclareAcronym{imc}{short=IMC,long=Internal Model Control,
	class= abbrev}
\DeclareAcronym{mpc}{short=MPC,long=Model Predictive Control,
	class= abbrev}
\DeclareAcronym{nmpc}{short=NMPC,long= Nonlinear Model Predictive Control,
	class= abbrev}  
\DeclareAcronym{lmpc}{short=LMPC,long= Linear Model Predictive Control,
	class= abbrev}
\DeclareAcronym{smpc}{short=SMPC,long= Stochastic Model Predictive Control,
	class= abbrev}
\DeclareAcronym{rmpc}{short=RMPC,long= Robust Model Predictive Control,
	class= abbrev}   

\DeclareAcronym{mrac}{short=MRAC,long=Model References Adaptive Control,
	class= abbrev}

\DeclareAcronym{dmc}{short=DMC,long=Dynamic Matrix Control,
	class= abbrev}
\DeclareAcronym{pfc}{short=PFC,long=Predictive Functional Control,
	class= abbrev}
\DeclareAcronym{gpc}{short=GPC,long=Generalized Predictive Control,
	class= abbrev}

\DeclareAcronym{mse}{short=MSE,long=Mean Squared Error,
	class= abbrev}
\DeclareAcronym{mae}{short=MAE,long=Mean Absolute Error,
	class= abbrev}
\DeclareAcronym{se}{short=SE,long=Standard Error,
	class= abbrev}
\DeclareAcronym{smse}{short=SMSE,long=Standardized Mean Squared Error,
	class= abbrev}

\DeclareAcronym{te}{short=TE,long=Tennessee Eastman,
	class= abbrev}
\DeclareAcronym{vtol}{short=VTOL,long=Vertical Take Off and Landing,
	class= abbrev}
\DeclareAcronym{dof}{short=DOF,long=Degree-of-Freedom,
	class= abbrev}

\DeclareAcronym{pid}{short=PID,long=Proportional-Integral-Derivative,
	class= abbrev}
\DeclareAcronym{lqr}{short=LQR,long=Linear-Quadratic Regulator,
	class= abbrev}
\DeclareAcronym{lmi}{short=LMI,long=Linear Matrix Inequality,
	class= abbrev}

\DeclareAcronym{mle}{short=MLE,long=Maximum Likelihood Estimation,
	class= abbrev}
\DeclareAcronym{ls}{short=LS,long=Least Square,
	class= abbrev}

\DeclareAcronym{ppd}{short=PPD,long=Pseudo-Partial Derivative,
	class= abbrev}

\DeclareAcronym{mfac}{short=MFAC,long=Model-Free Adaptive Control,
	class= abbrev}
\DeclareAcronym{cfdl}{short=CFDL,long=Compact Form Dynamic Linearization,
	class= abbrev}
\DeclareAcronym{pfdl}{short=PFDL,long=Partial Form Dynamic Linearization,
	class= abbrev}

\DeclareAcronym{ddc}{short=DDC,long=Data Driven Control,
	class= abbrev}

\DeclareAcronym{dp}{short=DP,long=Dynamic Programming,
	class= abbrev}
\DeclareAcronym{adp}{short=ADP,long=Approximate Dynamic Programming,
	class= abbrev}

\DeclareAcronym{mdp}{short=MDP,long=Markovian Decision Process,
	class= abbrev}

\DeclareAcronym{pilco}{short=PILCO,long= Probabilistic Inference for Learning Control,
	class= abbrev}

\DeclareAcronym{lp}{short=LP,long= Linear Programming,
	class= abbrev}  
\DeclareAcronym{nlp}{short=NLP,long= Nonlinear Programming,
	class= abbrev}  

\DeclareAcronym{wrt}{short= $\mathtt{WRT}$,long= with respect to,
	class= abbrev}

\DeclareAcronym{kkt}{short= KKT,long= Karush-Kahn-Tucker,
	class= abbrev}

\DeclareAcronym{qp}{short= QP,long= Quadratic Programming,
	class= abbrev}   

\DeclareAcronym{sqp}{short= SQP,long= Sequential Quadratic Programming,
	class= abbrev}

\DeclareAcronym{fpsqp}{short= FP-SQP,long= Feasibility-Perturbed Sequential Quadratic Programming,
	class= abbrev}   

\DeclareAcronym{mfcq}{short= MFCQ,long= Mangasarian-Fromovitz Constraint Qualification,
	class= abbrev}   

\DeclareAcronym{licq}{short= LICQ,long= Linear Independence Constraint Qualification,
	class= abbrev}

\DeclareAcronym{iae}{short= IAE,long= Integral Absolute Error,
	class= abbrev}    

\DeclareAcronym{bfgs}{short= BFGS,long= Broyden-Fletcher-Goldfarb-Shanno,
	class= abbrev}

\everymath{\displaystyle}
\newcommand{\colourred}{black}

\begin{document}%
	
\setcounter{page}{1}
	
\LRH{G. Cao et~al.}
	
\RRH{Enhanced PSOs for MIMO System Modelling using CGP Models}
	
\VOL{x}
	
\ISSUE{x}
	
\PUBYEAR{xxxx}

\JOURNALNAME{\TEN{}}%
	
%\BottomCatch
	
%\CLline
\PUBYEAR{201X}
	
%\subtitle{}
	
\title{Enhanced Particle Swarm Optimization Algorithms for Multiple-Input Multiple-Output System Modelling using Convolved Gaussian Process Models}

\authorA{Gang Cao}
\affA{School of Engineering and Advanced Technology\\ Massey University\\ Auckland, New Zealand\\
E-mail: g.cao@massey.ac.nz}

\authorB{Edmund M-K Lai}
\affB{Department of Information Technology and Software Engineering,\\ Auckland University of Technology, \\Auckland, New Zealand\\
	E-mail: edmund.lai@aut.ac.nz}
\authorC{Fakhrul Alam}
\affC{School of Engineering and Advanced Technology\\ Massey University\\ Auckland, New Zealand\\
	E-mail: f.alam@massey.ac.nz}

\begin{abstract}
\ac{cgp} is able to capture the correlations not only between inputs and outputs but also among the outputs.
This allows a superior performance of using \ac{cgp} than standard~\ac{gp} in the modelling of~\ac{mimo} systems when observations are missing for some of outputs.
Similar to standard~\ac{gp},
a key issue of \ac{cgp} is the learning of hyperparameters from a set of input-output observations.
It typically performed by maximizing the~\ac{ll} function which leads to an unconstrained nonlinear and non-convex optimization problem.
Algorithms such as~\ac{cg} or~\ac{bfgs} are commonly used but they often get stuck in local optima, especially for CGP where there are more hyperparameters.
In addition, the~\ac{ll} value is not a reliable indicator for judging the
quality intermediate models in the optimization process.
In this paper, we propose to use enhanced~\ac{pso} algorithms to solve this problem by minimizing the model output error instead.
This optimization criterion enables the quality of intermediate solutions to be directly observable during the optimization process.
Two enhancements to the standard~\ac{pso} algorithm which make use of gradient information and the multi-start technique are proposed.
Simulation results on the modelling of both linear and nonlinear systems demonstrate the effectiveness of minimizing the model output error to learn hyperparameters and the performance of using enhanced algorithms.
\end{abstract}
	
\KEYWORD{Enhanced PSO; Convolved Gaussian Process Models; Hyperparameters Learning}

\maketitle

\acresetall
%\SetAlFnt{\footnotesize}  % set small font size in the algorithm
\SetAlFnt{\small}

\newcommand{\newargmin}{\mathop{\mathrm{argmin}}} 
\newcommand{\newargmax}{\mathop{\mathrm{argmax}}} 
%  tikz 
\newcommand{\scaledegree}{0.8}
\definecolor{mygray}{RGB}{194,204,208}

\newcommand*{\params}{\varTheta}
\newcommand*{\hyperparas}{\bm{\theta}}
\newcommand*{\bighyperparas}{\bm{\Theta}}
\newcommand*{\meanvalue}{\bm{\mu}}
\newcommand*{\varvalue}{\bm{\Sigma}}

\newcommand{\singlefigWidth}{0.875}
\newcommand{\multifigWidth}{0.415}
\newcommand{\triplefigWidth}{0.315}
\newcommand{\tabcolsepwidth}{10pt}
\newcommand{\longtabcolsepwidth}{20pt}

\SetKwRepeat{Do}{do}{while}%
\newcommand{\nosemic}{\renewcommand{\@endalgocfline}{\relax}}% Drop semi-colon ;
\newcommand{\dosemic}{\renewcommand{\@endalgocfline}{\algocf@endline}}% Reinstate semi-colon ;
\newcommand{\pushline}{\Indp}% Indent
\newcommand{\popline}{\Indm\dosemic}% Undent
\let\oldnl\nl% Store \nl in \oldnl
\newcommand{\nonl}{\renewcommand{\nl}{\let\nl\oldnl}}% Remove line number for one line
\renewcommand{\KwIn}{\textbf{Initialization}}

% newline of table content 
\newcommand{\tabincell}[2]{\begin{tabular}{@{}#1@{}}#2\end{tabular}}

\section{Introduction}

\ac{gp} modelling is a non-parametric data-driven technique based on Bayesian theory.
A major advantage of~\ac{gp} models,
compared with parametric data-driven models such as~\ac{ann} and~\ac{fm}, 
is that the accuracy of the predicted outputs 
can be naturally measured through the variances that are computed as part of the modelling process.
Another advantage is that~\ac{gp} models generally require fewer parameters~\parencite{IP-Kocijan-2011}.
These parameters, also known as hyperparameters, are estimated through a 
learning process using the measured input-output data of the system. 
\ac{gp} models have found many applications in science and engineering~\parencite{A-BailerJone-SteelModellingUsingGPs-1999,A-Avzman-Application-2007,A-Wang-HumanMotionUsingGPs-2008,A-Gregorvcivc-ModellingNonlinear-2009,A-Yu-DignosisAndMonitoring-2012}.

A standard~\ac{gp} model can be applied to a~\ac{miso} system.
For systems with multiple outputs,
one can use a separate~\ac{gp} model for each output.
This approach is referred to as~\ac{igp} modelling.
Its disadvantage is that since the \ac{gp} models are independent of each other, any correlations between outputs will not be modelled~\parencite{IP-Boyle-Dependent-2005,IP-Alvarez-Sparse-2009,IP-Gang-2014a}.
An alternative way is to use \ac{cgp} models~\parencite{IP-Alvarez-Sparse-2009},
which are able to model not only the relationships between inputs and outputs but also correlations among all outputs.
The importance of modelling this correlation becomes apparent when there are missing output data~\parencite{IP-Gang-2014a}.

The hyperparameters of the~\ac{cgp} model can be estimated
by maximizing a~\ac{ll} function. 
This maximization is typically performed by using gradient based solutions, such as~\ac{cg} and~\ac{bfgs} algorithms.
The algorithms are usually required to restart many times with different initial values to overcome the issue of getting stuck in local optima caused by the sensitiveness to initial values. 
Evolutionary algorithms, such as standard~\ac{pso}, have been used as an alternative approach to learn the hyperparameters of~\ac{gp}~\parencite{IP-Zhu-Particle-2010,IP-Petelin-EvolvingGP-2011} and~\ac{cgp} model~\parencite{IP-Gang-2014a} due to they typically perform better than gradient based methods~\parencite{A-Noel-GradientbasedPSO-2012}.
However, the issues of poor global search ability caused by poor initialization and slow convergence due to poor local search ability remained in the existing works due to the use of standard~\ac{pso}.
In addition,
a physically meaningful and reliable indicator of intermediate models' quality is preferred than the use of~\ac{ll} values.

In view of these shortcomings, we propose three enhanced \ac{pso} algorithms to solve the optimization problem of minimizing~\ac{mse} values of model outputs.
The first one is called multi-start~\ac{pso} where the standard~\ac{pso} is restarted several times to diversify the particles.
The second one is the gradient-based~\ac{pso} which makes use of gradient information to achieve faster convergence.
The last one is a hybrid of these two methods that provides good particle diversity and faster convergence.
These three algorithms are studied through the modelling of~\ac{mimo}~\ac{ltv} and~\ac{nltv} systems.
Furthermore, the use of~\ac{mse} as fitness function provides us a direct and reliable indication of current solutions during the optimization process.

The rest of this article is organized as follows. 
Section~\ref{sec:cgps} provides a brief overview of the~\ac{cgp} modelling technique.
In Section~\ref{sec:hyperparamLearning}, 
we reviewed the maximizing~\ac{ll} function problem for learning~\ac{cgp} model' hyperparameters,
and defined the problem of minimizing~\ac{mse} of model outputs.
The standard~\ac{pso} as well as three enhanced algorithms for the problems are introduced in Section~\ref{sec:psolearning}.
Simulation results comparing the proposed algorithms to standard~\ac{pso} and \ac{cg} approaches are presented and discussed in Section~\ref{sec:simulation}.
Finally, Section~\ref{sec:conclude} concludes the article.

\section{Convolved Gaussian Process Models}
\label{sec:cgps}

Consider a system with $n$ inputs $\mathbf{x} \in \mathbb{R}^n$ and $m$ outputs $ \mathbf{y(x)}\in\mathbb{R}^m$ again. 
In the~\ac{cgp}, each output $\mathbf{y}_d(\mathbf{x})$ is modelled by,
\begin{equation}
\mathbf{y}_d(\mathbf{x}) = f_d(\mathbf{x}) + \epsilon_d(\mathbf{x})
\end{equation}
where $d=1,2,\ldots,m$ and $\epsilon_d(\mathbf{x})$ denotes an independent Gaussian white noise.
The function $f_d(\mathbf{x})$ typically is defined by a linear convolution of a smoothing kernel $H_d(\mathbf{x})$ and a latent function $u(\mathbf{x})$,
\begin{equation}
f_d(\mathbf{x}) 
= \int{H_{d}(\mathbf{x-\tau})u(\mathbf{\tau})d{\mathbf{\tau}}}
\end{equation}
The correlation between outputs is derived from the latent function $u(\mathbf{x})$ which has effects on all output functions. 
This latent function can be any appropriate random processes.
If a Gaussian white noise is used, then resulting in a~\ac{dgp} model. 
In the~\ac{cgp}, a wide range of latent functions are proposed to match the modelling requirements for different physical or dynamical systems~\parencite{PHD-Alvarez-Convolved-2011}.

In addition, the~\ac{cgp} models allow using more than one type of latent function.
Assuming $Q$ groups of latent functions are considered,
where for the $q^{th}$ group, it has $R_q$ smoothing kernels. 
Thus the $d^{th}$ output function can be rewritten by,
\begin{equation}
f_d(\mathbf{x}) 
=\sum_{q=1}^{Q}\sum_{k=1}^{R_{q}}\int{H_{d,q}^k(\mathbf{x-\tau})u_q^k(\mathbf{\tau})d{\mathbf{\tau}}}
\label{eqn:OutFucwithMultiLatent}
\end{equation}

Then, the covariance between different outputs $\mathbf{y}_d(\mathbf{x})$ and $\mathbf{y}_{d'}(\mathbf{x'})$ can be obtained by,
\begin{equation}
\begin{aligned}
\mathbf{K}_{\mathbf{y}_d,\mathbf{y}_{d'}}(\mathbf{x,x'})=&\textit{Cov}\left[\mathbf{y}_d(\mathbf{x}), \mathbf{y}_{d'}(\mathbf{x}')\right]\\
=&\textit{Cov}\left[f_d(\mathbf{x}),f_{d'}(\mathbf{x'})\right]
+\textit{Cov}\left[\epsilon_d(\mathbf{x}),\epsilon_{d'}(\mathbf{x'})\right]\delta_{d,d'} 
\end{aligned}
\label{eqn:cgp-output-covariance}
\end{equation}
where $\delta_{d,d'}$ is a Kronecker delta function thus $\textit{Cov}\left[\epsilon_d(\mathbf{x}),\epsilon_{d'}(\mathbf{x'})\right]\delta_{d,d'}$ will lead to a diagonal matrix of noise variance $\left\lbrace \sigma_d^2\right\rbrace_{d=1}^m$ if it is assumed that $\epsilon_d(\mathbf{x})~\sim\mathcal{N}(0,\sigma_d^2)$,
and the cross-covariance between 
$f_d(\mathbf{x})$ and $f_{d'}(\mathbf{x'})$ is given by,
\begin{equation}
\begin{aligned}
&\mathbf{K}_{f_d,f_{d'}}(\mathbf{x,x'}) =
\textit{Cov}\left[f_d(\mathbf{x}) ,f_{d'}(\mathbf{x}')\right]\\
&= \textit{E}\left[ 
\sum_{q=1}^{Q}\sum_{k=1}^{R_{q}}\int{H_{d,q}^k(\mathbf{x-\tau})u_q^k(\mathbf{\tau})d{\mathbf{\tau}}} 
\sum_{q=1}^{Q}\sum_{k=1}^{R_{q}}\int{H_{d',q}^k(\mathbf{x'-\tau'})u_q^k(\mathbf{\tau'})d{\mathbf{\tau'}}}
\right] \\
&=\sum_{q=1}^{Q}\sum_{k=1}^{R_{q}}k_q(\mathbf{\tau,\tau'})\int{H_{d,q}^k(\mathbf{x-\tau})H_{d',q}^k(\mathbf{x'-\tau})d{\mathbf{\tau}}}
\end{aligned}
\label{eqn:cgp-latent-covariance}
\end{equation}
Data-driven modelling using~\ac{cgp} basically involves obtaining the appropriate smoothing kernels and latent functions that reflect the covariance between outputs. 

As given in (\ref{eqn:OutFucwithMultiLatent}), the output function is a linear combination of independent random functions.
Thus, if these functions are Gaussian processes, then $f_{d}(\mathbf{x})$ will also be a Gaussian process.
In this case, the smoothing kernels can be expressed by,
\begin{equation}
H_{d,q}^k(\gamma) = \frac{\nu_{d,q}^k\left| {\mathbf{P}_{d,q}^k}\right|^{1/2} }{(2\pi)^{M/2}}\exp\left[ -\frac{1}{2}(\gamma)^T\mathbf{P}_{d,q}^k(\gamma)\right]
\label{eqn:smoothing}
\end{equation}
where $\nu_{d,q}^k$ is a length-scale coefficient, 
$\mathbf{P}_{d,q}^k$ is an $n\times n$ precision matrix of the smoothing kernel. 
To simplify the model further, it is assumed that the covariances of latent functions $k_q(\eta)$ in every group are all same Gaussian,
\begin{equation}
k_q(\eta)= \frac{\upsilon_q\left| {\mathbf{P}_q}\right|^{1/2} }{(2\pi)^{M/2}}\exp\left[ -\frac{1}{2}(\eta)^T\mathbf{P}_q(\eta)\right]
\label{eqn:covlatentfn}
\end{equation}
where $\upsilon_q$ is the length-scale coefficient and $\mathbf{P}_q$ is another $n\times n$ precision matrix.

To simplify the discussion again, it is assumed that $R_{q}=1$ for all $Q$ groups of latent functions. 
In addition, the precision matrices of the smoothing kernels are assumed to be the same for each group of latent functions. 
As a result, given the smoothing kernel (\ref{eqn:smoothing}) and latent function covariance (\ref{eqn:covlatentfn}), 
the covariance can be obtained by,
\begin{equation}
\textit{Cov}\left[f_d(\mathbf{x}) ,f_{d'}(\mathbf{x'})\right]
=\sum_{q=1}^{Q}\frac{\nu_{d,q}\nu_{d',q}\upsilon_q}{(2\pi)^{M/2}\left|\mathbf{P}\right|^{1/2}}
\exp\left[-\frac{1}{2}(\mathbf{x-x'})^T\mathbf{P}^{-1}(\mathbf{x-x'})\right]
\label{eqn:covariance} 
\end{equation}
where $\mathbf{P}=\mathbf{P_d^{-1}}+\mathbf{P_{d'}^{-1}}+\mathbf{P_q^{-1}}$. 
Note that this multiple-output covariance function maintains a Gaussian form, i.e. $\mathbf{K_{f_d,f_{d'}}(x,x')}\sim\mathcal{N}(\mathbf{x-x'}|0,\mathbf{P})$.

%It is assumed that the hyperparameters $\hyperparas=[\bm{\nu},\bm{\upsilon},\textbf{vec}(\mathbf{P})]^T$ are learnt.
Then similar to standard~\ac{gp} models, given a set of observations $\left\lbrace\mathbf{x}^j,\mathbf{y}_j \right\rbrace^{J_d}_{j=1}$, where $\sum_{d=1}^{m}J_d =N$,
a Gaussian distribution can be defined on the output functions by,
\begin{equation}
\mathbf{y(x)}\sim\mathcal{N}\left(\mathbf{\mu(x)},\mathbf{K_{y,y}(x,x')}\right) 
\end{equation}
where the output vector $\mathbf{y(x)}$ is given by,
\begin{equation}
\mathbf{y(x)}=\left[\mathbf{y}_1(\mathbf{x}),...,\mathbf{y}_m(\mathbf{x})\right]^T
\end{equation}
with the entries,
\begin{equation}
\mathbf{y}_d(\mathbf{x})=
\left[\mathbf{y}_d(\mathbf{x}^1),\mathbf{f_d}(\mathbf{x}^2),...,\mathbf{f_d}(\mathbf{x}^{J_d})\right]^T
\end{equation}
Without loss of generality, zero means are used.
In addition,
the covariance matrix $\mathbf{K_{y,y}(x,x')}\in\mathbb{R}^{N\times N}$ can be obtained by using (\ref{eqn:cgp-latent-covariance}) and (\ref{eqn:covariance}).
Usually, the computation of such a covariance matrix is computationally expensive.
Thus, some sparse approximations have been proposed to reduce the complexities of~\ac{cgp}~\parencite{IP-Alvarez-Sparse-2009}.
Then, the marginal likelihood can be defined by,
\begin{equation}
p(\mathbf{y|X},\hyperparas)\sim\mathcal{N}(\mathbf{y}|0,\mathbf{K}_{\mathbf{y,y}})
\label{eqn:priordistribution} 
\end{equation}
The joint distribution of observed $\mathbf{y}$ and the predicted outputs $\mathbf{y^*} = \left\lbrace y_1^*,\cdots,y_M^*\right\rbrace $ at new input $\mathbf{x^*}$ is thus still a Gaussian and is given by,
\begin{equation}
\left[                
\begin{array}{c}  
\mathbf{y}\\
\mathbf{y^*}\\
\end{array}
\right]\sim\mathcal{N}
\left(
\begin{array}{cc}
0,&\begin{array}{cc}
\mathbf{K_{y,y}}&\mathbf{K_{f,f^*}}\\
\mathbf{K_{f^*,f}}&\mathbf{K_{f^*,f^*}}
\end{array}
\end{array}
\right)             
\end{equation}
Finally, similar to standard~\ac{gp} models again,
the predictive distribution is a Gaussian,
\begin{equation}\label{eqn:cgp_preddistribution}
\mathbf{y^*|X,y,\hyperparas,x^*}\sim\mathcal{N}(\mu(\mathbf{x}^*),\sigma^2(\mathbf{x}^*))        
\end{equation}
where the mean $\mu(\mathbf{x}^*)$ and variance $\sigma^2(\mathbf{x}^*)$ functions are computed by,
\begin{eqnarray}\label{eqn:cgp_mean+var}
\begin{aligned}
\mu(\mathbf{x}^*)&=\mathbf{K_{f^*,f}}\mathbf{K_{y,y}^{-1}}\mathbf{y}
\label{eqn:preMean}\\
\sigma^2(\mathbf{x}^*)&=\mathbf{K_{f^*,f^*}}-\mathbf{K_{f^*,f}}\mathbf{K_{y,y}^{-1}}\mathbf{K_{f,f^*}}
\label{eqn:preVar}
\end{aligned}
\end{eqnarray}

\section{Hyperparameter Learning of CGP Models}
\label{sec:hyperparamLearning}

\subsection{Maximizing the Log-Likelihood Function}
When doing predictions using (\ref{eqn:cgp_mean+var}),
the covariance matrix $\mathbf{K}$ is required to be specified by a set of appropriate hyperparameters $\hyperparas$.
They are usually obtained by maximizing the log of marginal likelihood function.

In~\ac{cgp} models, the marginal likelihood is equal to the integral over a product of the likelihood function and~\ac{cgp} prior over the latent functions, 
both of which are Gaussian. 
Thus, the marginal likelihood is also Gaussian and defined by,
\begin{equation}
\begin{aligned}
p(\mathbf{y|X},\hyperparas) 
&= \int p(\mathbf{y|f,X},\hyperparas)
p(\mathbf{f}|\hyperparas)d\mathbf{f}\\
&= \frac{1}
{(2\pi)^{\frac{N}{2}}
	|\mathbf{K_{y,y}}|^{\frac{1}{2}}}
\exp\left(-\frac{1}{2}\mathbf{y}^T
\mathbf{K_{y,y}^{-1}}
\mathbf{y}
\right) 
\label{eqn:marginallikelihoodforGP}
\end{aligned}
\end{equation}
This marginal likelihood can be viewed as the likelihood of hyperparameters corrupted by noise so that we simply call likelihood function.
A good point estimate $\hat{\hyperparas}$ of hyperparameters can be subsequently obtained by maximizing this likelihood function. 
In practice,
we usually estimate the hyperparameters by maximizing the log likelihood function due to its less computation complexities.
The corresponding optimization problem can be subsequently defined as,
\begin{equation}\label{eqn:optpro_LL}
\hat{\hyperparas}=\newargmax_{\hyperparas} \log{p(\mathbf{y|X},\hyperparas)}
\end{equation}
where,
\begin{equation}\label{eqn:loglikelihood}
\log{p(\mathbf{y|X},\hyperparas)}=-\frac{1}{2}\mathbf{y}^T\mathbf{K_{y,y}^{-1}}\mathbf{y}-\frac{1}{2}\log{\left|\mathbf{K_{y,y}}\right|}-\frac{N}{2}\log{2\pi}
\end{equation}

The unconstrained optimization problem (\ref{eqn:optpro_LL}) is not easy to solve due to it is typically nonlinear and non-convex.
However, in~\ac{cgp} models,
the derivatives of log likelihood function with respective to (w.r.t.) the hyperparameters $\hyperparas$ are mathematically analytical and can be obtained by,
\begin{equation}
\frac{\partial}{\partial \hyperparas_l}\log{p(\mathbf{y|X},\hyperparas)}
=-\frac{1}{2}\mathbf{y}^T\mathbf{K_{y,y}^{-1}}\frac{\partial\mathbf{K}}{\partial\hyperparas_l}\mathbf{K_{y,y}^{-1}}\mathbf{y}-\frac{1}{2}\text{trace}(\mathbf{K_{y,y}^{-1}}\frac{\partial\mathbf{K}}{\partial\hyperparas_l})
\end{equation}
where $\hyperparas_l$ represents the $l^{\text{th}}$ entry of hyperparameters $\hyperparas$.

\subsection{Minimizing the MSE Function}

Equation~(\ref{eqn:loglikelihood}) is the natural choice as the objective function for the hyperparameter learning problem.
However, there are some issues involved which we shall illustrate with the modelling of a single output nonlinear dynamic system.
The system is described by the following difference equation:
\begin{equation}
y(k) = 0.893y(k-1)+0.037y^2(k-1)-0.05y(k-2) +0.157u(k-1)-0.05u(k-1)y(k-1)
\label{eqn:system}
\end{equation}
where $u(k)$ is the input and $y(k)$ is the output at time instant $k$.
$1000$ uniformly distributed input values are randomly generated within the range
$(-2,4)$ and the corresponding outputs are computed.
From these input-output data, $200$ are randomly chosen for training the model.
The hyperparameters of the \ac{cgp} model are learned by minimizing the negative of the \ac{ll} (NLL) function.
The quality of the resulting \ac{cgp} model is evaluated by computing the \ac{mse} of the outputs given by
\begin{equation}\label{eqn:mse}
MSE=\frac{1}{N}\sum_{i=1}^{N}\left(\mathbf{y}_i-\hat{\mathbf{y}}_i(\hyperparas)\right)^2
\end{equation}
using a different set of $50$ values.
Here,  $N$ is the number of test data, 
$\mathbf{y}_i$ are the $i^{\text{th}}$ observed output values, and
$\hat{\mathbf{y}}_i$ is corresponding mean value of the predictive distribution obtained by (\ref{eqn:preMean}) given the hyperparameters $\theta$.

\begin{table}
	\centering
	\caption{NLL and MSE values of two \ac{cgp} models of system described by~(\ref{eqn:system}). }
	\setlength{\tabcolsep}{\tabcolsepwidth}
	\begin{tabular}{c|c|c}
		\hline
		& Model 1 & Model 2 \\
		\hline
		NLL & $\approx 51$ & $\approx 242269$ \\
		\hline
		MSE & 0.5313  & 0.0101 \\
		\hline
	\end{tabular}
	\label{tb:nllVSmse}
\end{table}

Table~\ref{tb:nllVSmse} shows two different \ac{cgp} models that results from limiting the search range
of the hyperparameters to $[0,100]$ for Model 1 and $[0,1]$ for Model 2.  
From the MSE values, it is clear that Model 2 is able to predict the outputs more accurately compared with Model 1.
However, the NLL value of Model 1 is much smaller than Model 2.
If the NLL function is the objective function for minimization, one may conclude that Model 1 is the better model.
Thus one cannot use the NLL (and hence the \ac{ll}) values to accurately gauge the quality of the
intermediate models obtained during the optimization process.

We therefore propose to minimize the~\ac{mse} function (\ref{eqn:mse}) to learn~\ac{cgp}'s hyperparameters by,
\begin{equation}\label{eqn:optpro_MSE}
\hat{\hyperparas}=\newargmin_{\hyperparas}\frac{1}{N}\sum_{i=1}^{N}\left(\mathbf{y}_i-\hat{\mathbf{y}}_i(\theta)\right)^2
\end{equation}
In addition,
the following derivatives of~\ac{mse} of outputs w.r.t. hyperparameters can be used to accelerate the optimization process,
\begin{equation}
\frac{\partial}{\partial \hyperparas_l}MSE =-\frac{2}{N}\sum_{i=1}^{N}\left\lbrace \left(\mathbf{y}_i-\hat{\mathbf{y}}_i(\hyperparas)\right)\frac{\partial\hat{\mathbf{y}}_i(\hyperparas)}{\partial\hyperparas}\right\rbrace
\end{equation}
with
\begin{equation}
\frac{\partial\hat{\mathbf{y}}_i(\hyperparas)}{\partial\hyperparas}
=\frac{\partial \mathbf{K_{f^*,f}}}{\partial \hyperparas}\mathbf{K_{y,y}^{-1}}\mathbf{y}-\mathbf{K_{f^*,f}}\mathbf{K_{y,y}^{-1}}\frac{\partial \mathbf{K_{y,y}}}{\partial \hyperparas}\mathbf{K_{y,y}^{-1}}\mathbf{y}
\end{equation}
where the computation of $\frac{\partial \mathbf{K_{f^*,f}}}{\partial \hyperparas}$ and $\frac{\partial \mathbf{K_{y,y}}}{\partial \hyperparas}$ can be found in~\parencite{B-GPMLbook-2006,A-Alvarez-CGPs-2011}.
This technique is in fact widely known as the least-square approach in the literature.
In addition, from the viewpoint of non-Bayesian learning, 
minimizing the~\ac{mse} is approximately equivalent to maximizing the~\ac{ll}.
The proof of equivalence between these two learning strategies can be found in~\parencite{A-Myung-TutorialMLE-2003}.

\section{Enhanced PSO Algorithms}
\label{sec:psolearning}

%\ac{pso} is an evolutionary computation technique inspired by the social behaviour of organisms~\parencite{IP-Pso-1995}.

%Many particles are initialized simultaneously and each one represents a solution to the problem.

%Associated with each particle is its position in the solution space and its velocity with which it is moved to a new position.

%A fitness function is used to evaluate each particle and only particles that are fit enough survive in the competition. 

%After some iterations, the particles would have explored the solution space sufficiently to arrive at the optimal or near optimal solution.

In~\parencite{IP-Zhu-Particle-2010,IP-Petelin-EvolvingGP-2011,IP-Gang-2014a},
the standard~\ac{pso} algorithm has been proven superior to gradient based~\ac{cg} and~\ac{bfgs} approaches in terms of accuracy and efficiency for the optimization problems (\ref{eqn:optpro_LL}) and (\ref{eqn:optpro_MSE}).
However,
poor initializations can lead to poor global search ability, and they exhibit slow convergence due to poor local search ability.
In this section,
three enhancements are proposed to address these issues.

%
% Standard
%
\subsection{Standard PSO}
\label{subsec:spso}

\begin{algorithm}[!t]
	\setstretch{0.1}
	\textbf{Initialization}\\
	\nonl PSO parameters: $N_p, c_1, c_2, \lambda_1, \lambda_2, \omega_\text{start}, \omega_\text{end},k, T_{\text{max}}$ and $\xi$\\
	\nonl Randomly generated $\hyperparas$;\\
	\BlankLine
	\While{$t<T_{\text{max}}$}{
		\eIf{$f(\mathbf{G})\leq\xi$}
		{End;}
		{\For{$i=1$ to $N_p$}
			{\For{$d=1$ to $D$}
				{Update $v_i^d(t)$ by using (\ref{eqn:standardPSO-v});\\
					Update $x_i^d(t)$ by using (\ref{eqn:standardPSO-x});\\}
				Update $\mathbf{P}_i$ and $V_i^{\text{pbest}}(t)$ by using (\ref{eqn:pbest});\\
				Update $\mathbf{G}$ and $V^{\text{gbest}}(t)$ by using (\ref{eqn:gbest});\\
			}
		}
		s $t=t+1$;\\
	}
	\KwOut{Optimized particle $\hyperparas_{\text{opt}}$.}
	\BlankLine
	\caption{Standard PSO based Hyperparameter Learning}
	\label{alg:spso}
\end{algorithm}

We shall first outline the standard~\ac{pso} algorithm for the hyperparameter learning of~\ac{cgp} models.
Let there be a population of $N_p$ particles,
each of which, denoted by $\mathbf{x}_i=[x_i^1,\cdots,x_i^D]_{i=1,\cdots,N_p}^T\in\mathbb{R}^D$,
represents a potential solution to the problem (\ref{eqn:optpro_LL}) or (\ref{eqn:optpro_MSE}).
Each particle also records its best position as $\mathbf{P}_i=[p_i^1,\cdots,p_i^D]^T$ and its best fitness value $V^{\text{pbest}}_i=f(\mathbf{P}_i)$,
where $f(\cdot)$ denotes the fitness function and could be (\ref{eqn:loglikelihood}) or (\ref{eqn:mse}).
In addition,
the best position of all $N_p$ particles is denoted by $\mathbf{G}=[g^1,\cdots,g^D]^T$ and the corresponding best fitness value is denoted by $V^{\text{gbest}}=f(\mathbf{G})$.
In the iteration $t+1$,
the velocity of $i^{\text{th}}$ particle, given by $\mathbf{v}_i=[v_i^1,\cdots,v_i^D]^T$, along $d^{\text{th}}$ dimension is updated according to the following rule,
\begin{equation}
v_i^d(t+1)=\omega(t)v_i^d(t)+c_1\lambda_1\left(p_i^d(t)-x_i^d(t)\right)+c_2\lambda_1\left(g^d(t)-x_i^d(t)\right)
\label{eqn:standardPSO-v}
\end{equation}
where $c_1$ and $c_2$ are two acceleration factors,
$\lambda_1$ and $\lambda_2$ are two random values between $[0,1]$,
$\omega(t)$ represents an inertia factor.

In general, a \ac{pso} algorithm consists of two search phases, known as ``exploration" and ``exploitation" respectively.
They are governed by the inertia factor $\omega(t)$.
The use of a larger value of $\omega(t)$ allows the particle to explore larger areas of the search space during the exploration phase.
Meanwhile, 
a smaller value of $\omega(t)$ restricts the particle to a smaller region of the search space and allows the particle to converge to a local optimum in the exploitation phase.
Thus, the inertia factor is usually reduced with time step.
A commonly used $\omega(t)$ is defined by,
\begin{equation}
\omega(t)=\omega_{\text{end}}+(\omega_{\text{start}}-\omega{_\text{end}})\exp(-k\times(\frac{t}{T_{\text{max}}}))
\label{eqn:omega}
\end{equation}
where $\omega_{\text{start}}$ and $\omega_{\text{end}}$ are the pre-determined start and final values respectively,
$T_{\text{max}}$ denotes the maximum number of iterations.
and the rate of decrease is governed by the constant $k$. 

The new position of a particle can subsequently be obtained by,
\begin{equation}
x_i^d(t+1) = x_i^d(t) + v_i^d(t+1)
\label{eqn:standardPSO-x}
\end{equation}
For the minimization problem (\ref{eqn:optpro_MSE}),
the $\mathbf{P}_i$ and $V_i^{\text{pbest}}$ at $t+1$ iteration are updated according to the following rule,
\begin{eqnarray}\label{eqn:pbest}
\begin{aligned}
\label{eqn:pbest-p}\mathbf{P}_i(t+1)=&\left\lbrace
\begin{array}{ll}
\mathbf{x}_i(t+1) & f(\mathbf{x}_i(t+1)) \leq f(\mathbf{P}_i(t))\\
\mathbf{P}_i(t)& f(\mathbf{x}_i(t+1)) > f(\mathbf{P}_i(t))
\end{array}
\right.\\
\label{eqn:pbest-v}V_i^{\text{pbest}}(t+1)=&f(\mathbf{P}_i(t+1))
\end{aligned}
\end{eqnarray}
In addition,
the $\mathbf{G}$ and $V^{\text{gbest}}$ at $t+1$ iteration are updated by,
\begin{eqnarray}\label{eqn:gbest}
\begin{aligned}
\label{eqn:gbest-p}\mathbf{G}(t+1)
&=\newargmin\bigg\{ f(\mathbf{P}_1(t+1)),\cdots,f(\mathbf{P}_{N_p}(t+1)), f(\mathbf{G}(t))\bigg\}\\
\label{eqn:gbest-v}V^{\text{gbest}}(t+1)&=f(\mathbf{G}(t+1))
\end{aligned}
\end{eqnarray}
We can also use the rules (\ref{eqn:pbest}) and (\ref{eqn:gbest}) when the maximization problem (\ref{eqn:optpro_LL}) becomes the minimizing negative of~\ac{ll} function (\ref{eqn:loglikelihood}).

For our hyperparameter learning problem,
each particle is defined by
\begin{equation}\label{eqn:particle}
\hyperparas = \left\lbrace \hyperparas_{K1},...,\hyperparas_{KM},\hyperparas_{L1},...,\hyperparas_{LQ}\right\rbrace 
\end{equation}
where $\hyperparas_{Kd} = \left\lbrace \nu_{d1},...,\nu_{dQ},\mathbf{P_d}\right\rbrace_{d=1,\cdots,M}$ represents the hyperparameters of smoothing kernels (\ref{eqn:smoothing}), 
and $\hyperparas_{Lq} = \left\lbrace \upsilon_{q},\mathbf{P_q}\right\rbrace_{q=1,\ldots,Q}$ are the hyperparameters of latent functions (\ref{eqn:covlatentfn}).
The algorithm of standard~\ac{pso} based hyperparameter learning is presented in Algorithm~\ref{alg:spso}.

%
% Multi-start
%
\subsection{Multi-Start PSO}
\label{subsec:multistartPSO}

\begin{algorithm}[!t]
	\setstretch{0.1}
	\textbf{Initialization}\\
	\nonl PSO parameters: $N_p, c_1, c_2, \lambda_1, \lambda_2, \omega_\text{start}, \omega_\text{end},k, T_{\text{max}}$ and $\xi$\\
	\nonl Randomly generated $\hyperparas$;\\
	\nonl Multi-start PSO parameters: $\eta, N_G$, $N_\eta =0$;\\
	\BlankLine
	\While{$t<T_{\text{max}}$}{
		\eIf{$N_\eta = N_G$}
		{Randomly regenerated $\hyperparas$;\\
			$N_\eta =0$;\\}
		{\eIf{$f(\mathbf{G})\leq\xi$}
			{End;}
			{\For{$i=1$ to $N_p$}
				{\For{$d=1$ to $D$}
					{Update $v_i^d(t)$ by using (\ref{eqn:standardPSO-v});\\
						Update $x_i^d(t)$ by using (\ref{eqn:standardPSO-x});\\}
					Update $\mathbf{P}_i$ and $V_i^{\text{pbest}}(t)$ by using (\ref{eqn:pbest});\\
					Update $\mathbf{G}$ and $V^{\text{gbest}}(t)$ by using (\ref{eqn:gbest});\\
				}
				\eIf{$\left\|f(\mathbf{G}(t))-f(\mathbf{G}(t-1))\right\| \leq \eta$}
				{$N_\eta = N_\eta +1$;}
				{$N_\eta = 0$;}
			}
		}
		$t=t+1$;\\
	}
	\KwOut{Optimized particle $\hyperparas_{\text{opt}}$.}
	\BlankLine
	\caption{Multi-Start PSO based Hyperparameter Learning}
	\label{alg:mpso}
\end{algorithm}

In the ``exploration" stage of optimization process, we want the particles to explore as much of the search space as possible.
This can be achieved by setting the inertia factor $\omega(t)$ to a suitably large value which in turn
is determined by $\omega_{start}$ and $\omega_{end}$ in (\ref{eqn:omega}).
However, suitable values for these two constants are quite specific to each problem.
Another way to achieve this objective is to diversify the swarm by introducing new particles.
In this paper,
all particles will be reinitialized if the global best position $\mathbf{G}$ remains unchanged or slightly changed for a given number of iterations $N_G$.
This is referred as the multi-start~\ac{pso} algorithm.
One issue remained in the proposed algorithm is that the potentials of old particles may not be sufficiently exploited.
This issue can be ignored due to we care the global search ability more than local one in the ``exploration" stage.
In addition,
it has been proposed that only those particles that are trapped in a local optimum should be reinitialized~\parencite{A-An-ModifiedPSO-2010}.
However,
the rest of particles may still need to be reinitialized later.
Besides,
this approach requires checking the changes of multiple $f(\mathbf{P}_i)$.
The proposed algorithm is therefore simpler due to only the change of $f(\mathbf{G})$ is checked.
Algorithm~\ref{alg:mpso} describes the approach of learning~\ac{cgp} models' hyperparameters through using the multi-start~\ac{pso}.

\subsection{Gradient-based PSO}
\label{subsec:gradientPSO}

\begin{algorithm}[!t]
	\setstretch{0.1}
	\textbf{Initialization}\\
	\nonl PSO parameters: $N_p, c_1, c_2, \lambda_1, \lambda_2, \omega_\text{start}, \omega_\text{end},k, T_{\text{max}}$ and $\xi$\\
	\nonl Randomly generated $\hyperparas$;\\
	\nonl Gradient-based PSO parameters: $\eta, N_G$, $N_\eta =0$;\\
	\BlankLine
	\While{$t<T_{\text{max}}$}{
		\eIf{$N_\eta = N_G$}
		{Initializing CG parameters, $\hyperparas_0=\mathbf{G}$;\\
			Solving the problem (\ref{eqn:optpro_LL}) or (\ref{eqn:optpro_MSE}) to obtain $\hyperparas^*$;\\
			\eIf{$f(\hyperparas^*)\leq f(\mathbf{G}(t))$}
			{$\mathbf{G}(t+1)=\hyperparas^*$;\\}
			{$\mathbf{G}(t+1)=\mathbf{G}(t)$;\\}
			$N_\eta =0$;\\}
		{\eIf{$f(\mathbf{G})\leq\xi$}
			{End;}
			{\For{$i=1$ to $N_p$}
				{\For{$d=1$ to $D$}
					{Update $v_i^d(t)$ by using (\ref{eqn:standardPSO-v});\\
						Update $x_i^d(t)$ by using (\ref{eqn:standardPSO-x});\\}
					Update $\mathbf{P}_i$ and $V_i^{\text{pbest}}(t)$ by using (\ref{eqn:pbest});\\
					Update $\mathbf{G}$ and $V^{\text{gbest}}(t)$ by using (\ref{eqn:gbest});\\
				}
				\eIf{$\left\|f(\mathbf{G}(t))-f(\mathbf{G}(t-1))\right\| \leq \eta$}
				{$N_\eta = N_\eta +1$;}
				{$N_\eta = 0$;}
			}
		}
		$t=t+1$;\\
	}
	\KwOut{Optimized particle $\hyperparas_{\text{opt}}$.}
	\BlankLine
	\caption{Gradient-based PSO based Hyperparameter Learning}
	\label{alg:gpso}
\end{algorithm}

Standard~\ac{pso} also suffers from slow convergence during the ``exploitation" phase.
This issue can be solved through using the gradient/derivative information especially when approaching to the global or local optima.
In this paper,
a gradient-based~\ac{pso} is proposed for the hyperparameters learning problem by combining the standard~\ac{pso} and~\ac{cg} algorithm.
In particular,
the current global best position $\mathbf{G}$ will be exploited by solving the problem \ref{eqn:optpro_LL}) or (\ref{eqn:optpro_MSE}) by using the~\ac{cg} algorithm.
The obtained solution is subsequently used to replace the current global position in the~\ac{pso} algorithm if it produces a better fitness value.
Compared with the existing work in~\parencite{A-Noel-GradientbasedPSO-2012} where all particles are exploited by using a gradient-based method,
the proposed algorithm only conducts gradient-based search on the current global best position if its fitness value remains unchanged or slightly changed for a specified number of iterations $N_G$.
The computational burden of using proposed algorithm is essentially reduced.
The gradient based~\ac{pso} for the hyperparameter learning of~\ac{cgp} models is given in Algorithm~\ref{alg:gpso}.

\subsection{Hybrid PSO}
\label{subsec:hybridPSO}

\begin{algorithm}[!t]
	\setstretch{0.1}
	\textbf{Initialization}\\
	\nonl PSO parameters: $N_p, c_1, c_2, \lambda_1, \lambda_2, \omega_\text{start}, \omega_\text{end},k, T_{\text{max}}$ and $\xi$\\
	\nonl Randomly generated $\hyperparas$;\\
	\nonl Hybrid PSO parameters: $\tau, \eta, N_G$, $N_\eta =0$;\\
	\BlankLine
	\While{$t<T_{\text{max}}$}{
		\eIf{$N_\eta = N_G$}
		{
			\eIf{$t\leq \tau\times T_{\text{max}}$}
			{Randomly regenerated $\hyperparas$;}
			{Initializing CG parameters, $\hyperparas_0=\mathbf{G}$;\\
				Solving the problem (\ref{eqn:optpro_LL}) or (\ref{eqn:optpro_MSE}) to obtain $\hyperparas^*$;\\
				\eIf{$f(\hyperparas^*)\leq f(\mathbf{G}(t))$}
				{$\mathbf{G}(t+1)=\hyperparas^*$;\\}
				{$\mathbf{G}(t+1)=\mathbf{G}(t)$;\\}
			}
			$N_\eta =0$;\\}
		{\eIf{$f(\mathbf{G})\leq\xi$}
			{End;}
			{\For{$i=1$ to $N_p$}
				{\For{$d=1$ to $D$}
					{Update $v_i^d(t)$ by using (\ref{eqn:standardPSO-v});\\
						Update $x_i^d(t)$ by using (\ref{eqn:standardPSO-x});\\}
					Update $\mathbf{P}_i$ and $V_i^{\text{pbest}}(t)$ by using (\ref{eqn:pbest});\\
					Update $\mathbf{G}$ and $V^{\text{gbest}}(t)$ by using (\ref{eqn:gbest});\\
				}
				\eIf{$\left\|f(\mathbf{G}(t))-f(\mathbf{G}(t-1))\right\| \leq \eta$}
				{$N_\eta = N_\eta +1$;}
				{$N_\eta = 0$;}
			}
		}
		$t=t+1$;\\
	}
	\KwOut{Optimized particle $\hyperparas_{\text{opt}}$.}
	\BlankLine
	\caption{Hybrid PSO based Hyperparameter Learning}
	\label{alg:hpso}
\end{algorithm}

The multi-start method in Section~\ref{subsec:multistartPSO} and the gradient-based method in Section~\ref{subsec:gradientPSO} 
can be combined in a single \ac{pso} algorithm so that both the ``exploration" and the ``exploitation" phases of the optimization process are enhanced.
This leads to the hybrid~\ac{pso} algorithm.
In particular,
the multi-start technique is first used such that the search space can be well covered.
When the number of iterations $N_G$ reaches a given proportion $\eta$ of maximum iteration number,
the optimization process is considered to have approached near global or local optima.
The algorithm subsequently switches to the use of gradient-based technique.
This allows a faster convergence rate due to the nature of using gradient-based solution compared to the use of rules (\ref{eqn:standardPSO-v}) and (\ref{eqn:standardPSO-x}).
The proposed hybrid~\ac{pso} is conceptually simple and allows to adjust the proportion $\eta$ to suit the problem.
The use of hybrid~\ac{pso} in the problem of~\ac{cgp} models' hyperparameter learning is given in Algorithm~\ref{alg:hpso}.

\section{Performance Evaluation}
\label{sec:simulation}

\textcolor{\colourred}{The performance of the proposed~\ac{pso} discussed in Section~\ref{sec:psolearning} for 
	\ac{cgp} hyperparameters learning is evaluated by computer simulation.
	We consider the modelling of non-trivial \ac{miso} and~\ac{mimo} systems in these numerical experiments.
	The results are compared with those obtained using the standard~\ac{cg} and \ac{bfgs}.
	In addition, results using both the \ac{nll} and the \ac{mse} as the fitness function are compared.}

\textcolor{\colourred}{All simulations are repeated $50$ times on a computer with a $3.40$GHz Intel$\circledR$ Core$^{\text{TM}}$ $2$ Duo CPU with $16$ GB RAM,
	using Matlab$\circledR$ version $8.1$.
	The average results of these $50$ simulation runs are shown here.
	Table~\ref{tb:cgpParameters} shows the key parameters of~\ac{cgp} and~\ac{pso} used in the simulations.}

\begin{table*}
	\centering
	\caption{Key parameters used in simulations}
	\begin{tabular}{ c | c | c }
		\hline
		\textbf{Symbol} & \textbf{Description} & \textbf{Quantity} \\
		\hline
		$N_p$ & PSO population & 20 \\  
		\hline
		$\mathbf{T}_\text{max}$ & Maximum Iterations & 500 \\  
		\hline
		$c_1$,$c_2$  & Acceleration Factors & 1.5\\  
		\hline
		$\omega_{\text{start}}$ & Start Inertial Factor & 0.4\\  
		\hline
		$\omega_{\text{end}}$   & End Inertial Factor & 0.9\\  
		\hline
		$k$    & Shape Control Factor & 0.8\\ 
		\hline
		$\left\| \Delta\xi\right\|$ & Minimum Fitness Variation & $10^{-5}$ \\
		\hline
		\multirow{3}{*}{\makecell{$\nu_{d,i}, \upsilon_{q}$ \\ $\alpha_i,\beta_j$}} & \multirow{3}{*}{\makecell{Coefficients Search Range\\ $\mathbf{P_d}$, $\mathbf{P_q}$ Elements Search Range}} & $[0,100]$ for LTV \\
		\cline{3-3}
		& & $[0,100]$ for NLTV with ``Step"\\ 
		\cline{3-3}
		& & $[0,1]$ for NLTV with ``Curve" \\ 
		\hline 
	\end{tabular}
	\label{tb:cgpParameters}
\end{table*}

\begin{table}[!t]
	\caption{The~\ac{mse} values of predicted outputs of the~\ac{cgp} models learned by 
		using standard PSO for system (\ref{eqn:system}).  $\text{PSO}/1$ uses \ac{mse} and $\text{PSO}/2$ uses\ac{nll} as fitness function.}
	\label{tb:SimofPopulation}
	\centering
	\begin{tabular}{c| c| c| c| c}
		\hline
		\multirow{2}{*}{$N_p$} &\multicolumn{2}{c|}{MSE} &\multicolumn{2}{c}{Time(seconds)} \\
		\cline{2-5}
		& $\text{PSO}/1$ & $\text{PSO}/2$ & $\text{PSO}/1$ & $\text{PSO}/2$ \\
		\hline
		10 & 0.2297 & 0.2355  & 9.21 & 9.95\\
		\hline
		20 & 0.0054 & 0.0047& 20.24 & 21.18\\
		\hline
		50 & 0.0022 & 0.0021& 25.21 & 27.69\\
		\hline
		100 & 0.0011 & 0.0012& 46.33 & 47.67\\
		\hline
	\end{tabular}
\end{table}

\begin{table}[!t]
	\caption{\ac{mse} of the predicted outputs for \ac{cgp} models learned by the proposed standard~\ac{pso} with~\ac{mse} fitness (PSO/2), \ac{cg} and~\ac{bfgs} in the two-output modelling problem, where $\mathbf{y}_2 = -\mathbf{y}_1$.}
	\label{tb:mimoCGPlearnResults-linear}
	\centering
	\begin{tabular}{c| c| c| c}
		\hline
		& PSO/2 & CG & BFGS \\
		\hline
		$y_1$ & 6.4587e-08 & 8.2713e-04 & 6.9378e-05\\
		\hline
		$y_2$ & 2.3900e-08 & 2.6176e-05 & 1.8735e-04\\
		\hline
	\end{tabular}
\end{table}

\begin{table}[!t]
	\caption{\ac{mse} of the predicted outputs for \ac{cgp} models learned by the proposed standard~\ac{pso} with~\ac{mse} fitness (PSO/2), \ac{cg} and~\ac{bfgs} in the two-output modelling problem, where $\mathbf{y}_2 = \textbf{exp}(\mathbf{y}_1)$.}
	\label{tb:mimoCGPlearnResults-nonlinear}
	\centering
	\begin{tabular}{c| c| c| c}
		\hline
		& PSO/2 & CG & BFGS \\
		\hline
		$y_1$ & 2.3141e-08 & 2.2108e-05 & 1.9892e-04\\
		\hline
		$y_2$ & 3.7233e-08 & 5.7204e-05 & 1.8949e-04\\
		\hline
	\end{tabular}
\end{table}

\subsection{Effects of Using MSE As Fitness Function}
\subsubsection{Single Output Modelling}
\label{subsubsec:singleoutput}
The system described by (\ref{eqn:system}) is used for modelling here.
Although this dynamical system has only $1$ input and $1$ output, 
the~\ac{cgp} modelling inputs will be $u(k-1)$, $y(k-1)$ and $y(k-2)$,
making it a $3$-input and $1$-output model.
Only a single output is used here for modelling to simplify the comparison.
In addition,
we randomly chose $1000$ inputs in $u\sim\mathcal{U}(-2,4)$ and apply them into the system.
This allows us to collect $1000$ observations including inputs, states and outputs.

\textcolor{\colourred}{$1000$ inputs for $u\sim\mathcal{U}(-2,4)$ are generated and applied to the system.
	This allows us collect $1000$ observations which includes the inputs, the states and the output.
	From this set of observations, $200$ training and $50$ test data are randomly selected.}

\textcolor{\colourred}{\ac{cgp} models are trained using the \ac{nll} and the \ac{mse} as fitness functions, denoted by PSO/1 and PSO/2 respectively,
	with the standard \ac{pso} algorithm.
	Table~\ref{tb:SimofPopulation} shows the \ac{mse} values using $50$ test samples on the resulting \ac{cgp} models, 
	for population sizes of $10$, $20$, $50$ and $100$.
	For all four population sizes,
	the~\ac{mse} of the predicted outputs for PSO/1 and PSO/2 are very close.
	This implies that using \ac{mse} produces models of similar quality as those obtained using \ac{nll}.
	Furthermore, PSO/1 and PSO/2 require similar amount of computation time.}

\textcolor{\colourred}{In both cases, a larger population size produces better quality models but require a longer computation time. 
	It seems that using a population size between $20$ to $50$ provides a good trade-off between model accuracy and computational efficiency. 
	Hence a population size of $20$ will be used for the rest of the simulations.}

\subsubsection{Two-output Modelling}
\label{subsubsec:twooutput}
Systems with multiple-outputs can be modelled in two different ways.
One is to use multiple single-output models and the other is to provide a single model for all outputs at the same time.
While the first approach is often simpler, the latter approach is able to capture correlation between outputs.
For example, a robot arm system with multiple degrees of freedom has multiple outputs that are strongly correlated.
Another example is the prediction of steel mechanical properties in~\parencite{IP-SteelMechPrediction-2010},
where the yield and tensile strength are predicted from the chemical compositions and grain size.
Note that these two outputs are highly correlated.

We shall continue to use the dynamical system in (\ref{eqn:system}).
Since it has only one output $y$ (denoted $y_1$ here), a second output $y_2$ will be created as a function of $y_1$.
Two such functions are considered, one linear and the other nonlinear, given by
$\mathbf{y}_2 = -\mathbf{y}_1$ and 
$\mathbf{y}_2 = \textbf{exp}(\mathbf{y}_1)$ respectively.
Two different sets of training data, each has $200$ samples, are selected from the $1000$ observations. 
The test data consists of $50$ samples which are different from the training samples. 
\textcolor{\colourred}{The performance of PSO/2 is compared that obtained by \ac{cg} and \ac{bfgs}.
	Note that \ac{cg} and \ac{bfgs} should be restarted $20\times 500$ times in order to provide a fair comparison to PSO/2.
	However this will result in much longer computation time than the PSO/2.
	In our simulations, \ac{cg} and~\ac{bfgs} are restarted $2000$ times so that the computation times of the three methods are comparable.}

\begin{figure*}[!t]
	\centering
	\subfigure[Linear -- $y_1$]{
		\label{fig:linear-y1}
		\centering
		\includegraphics[width=\multifigWidth\textwidth]{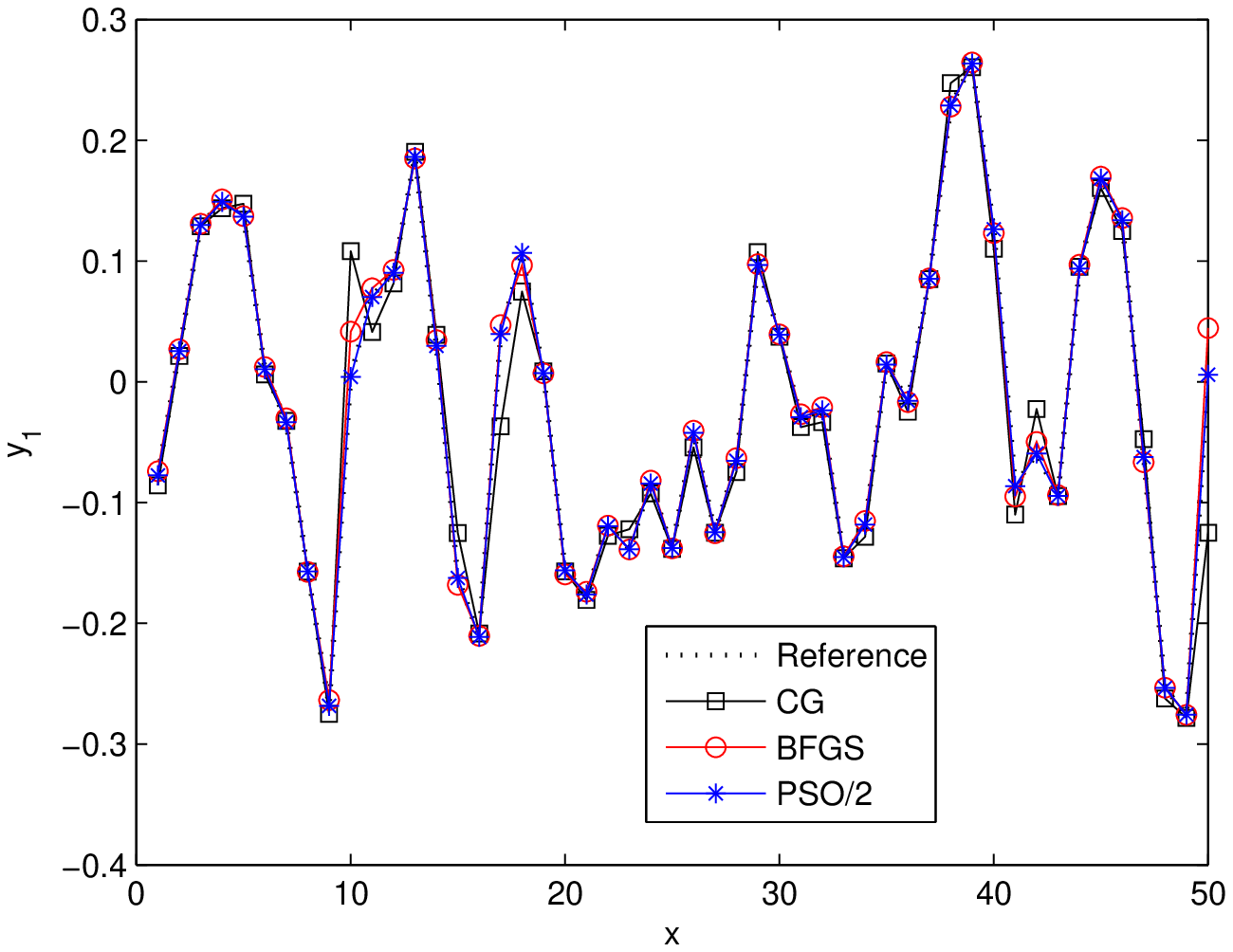}}
	\subfigure[Linear -- $y_2$]{
		\label{fig:linear-y2}
		\centering
		\includegraphics[width=\multifigWidth\textwidth]{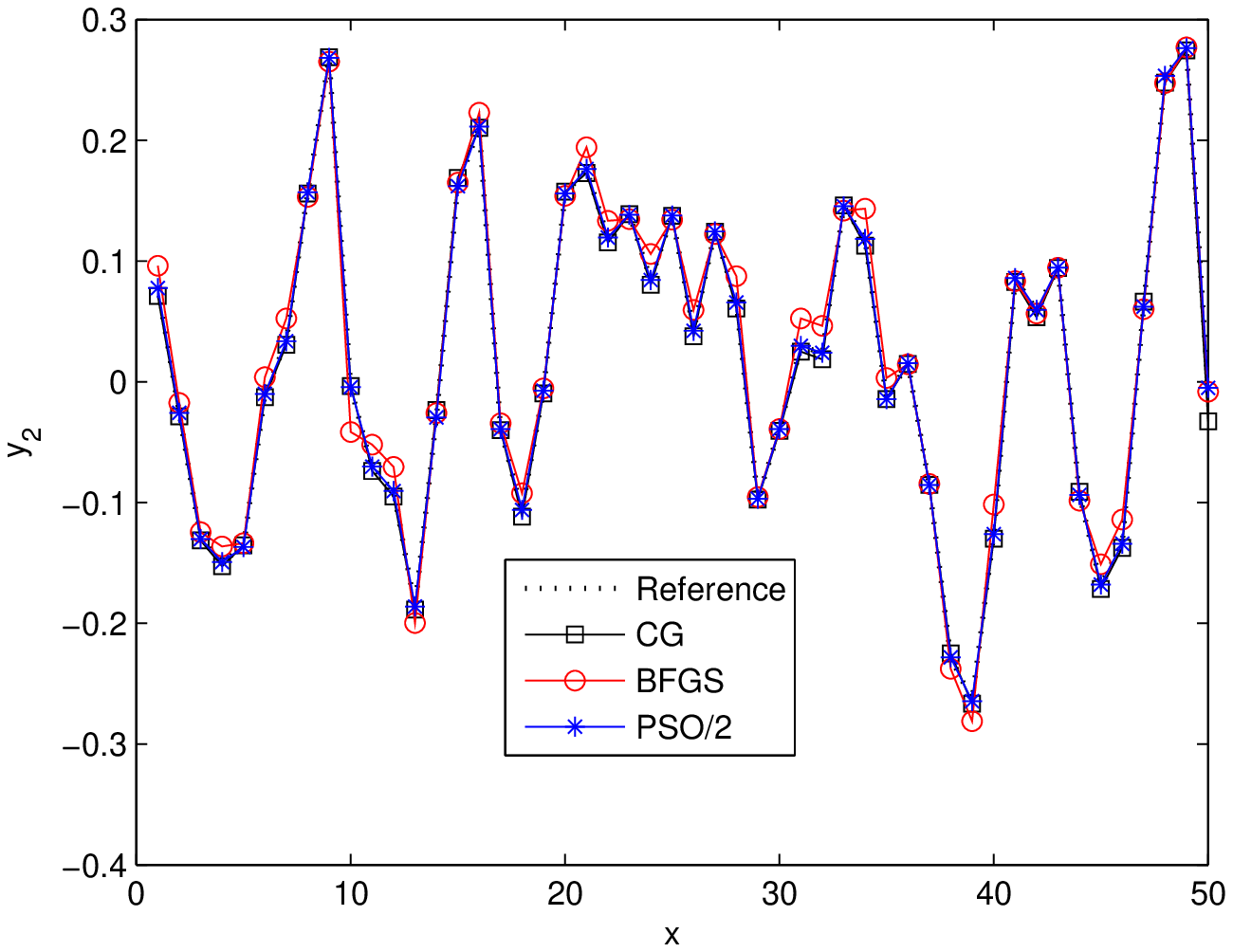}}\\
	\subfigure[Nonlinear -- $y_1$]{
		\label{fig:nonlinear-y1}
		\centering
		\includegraphics[width=\multifigWidth\textwidth]{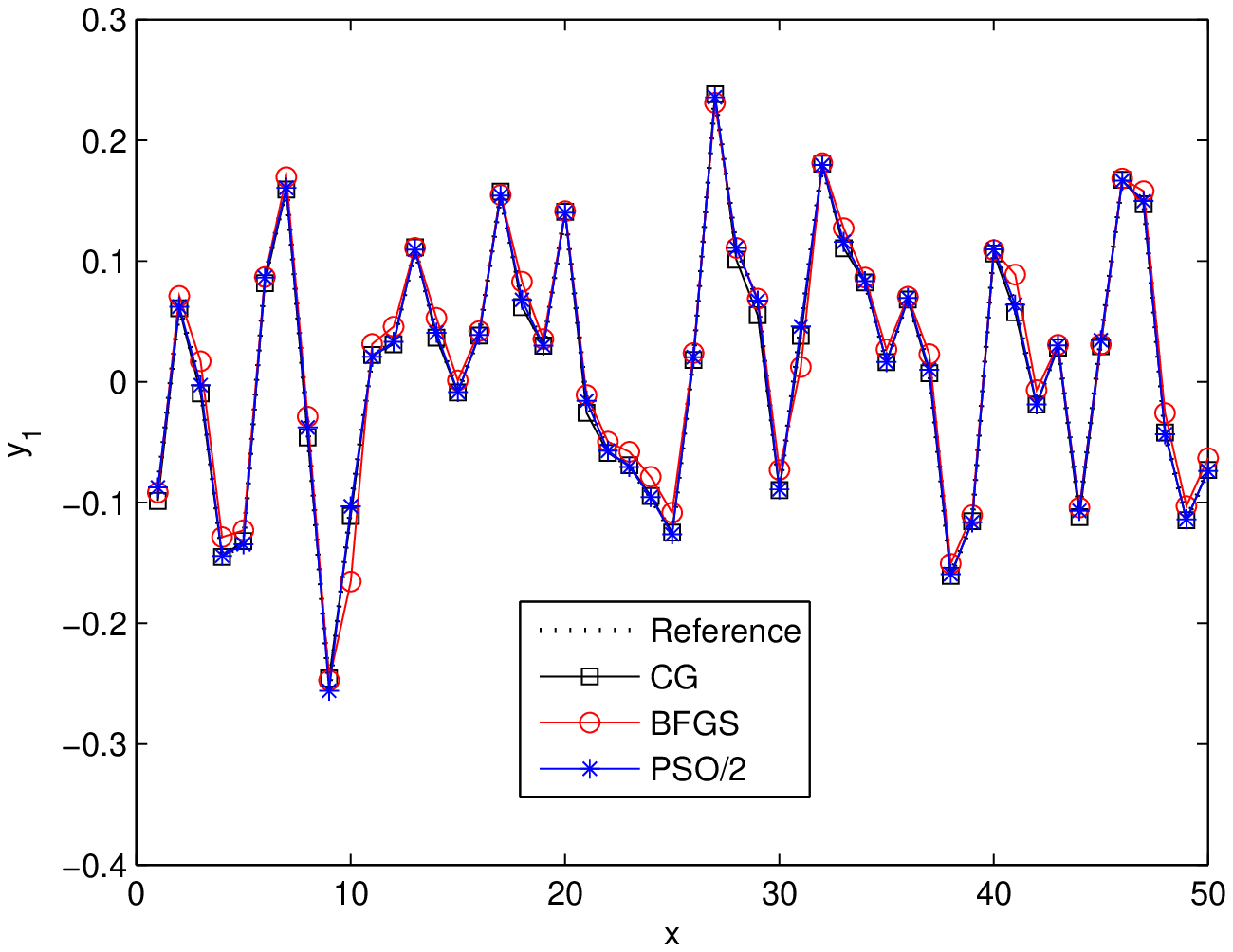}}
	\subfigure[Nonlinear -- $y_2$]{
		\label{fig:nonlinear-y2}
		\centering
		\includegraphics[width=\multifigWidth\textwidth]{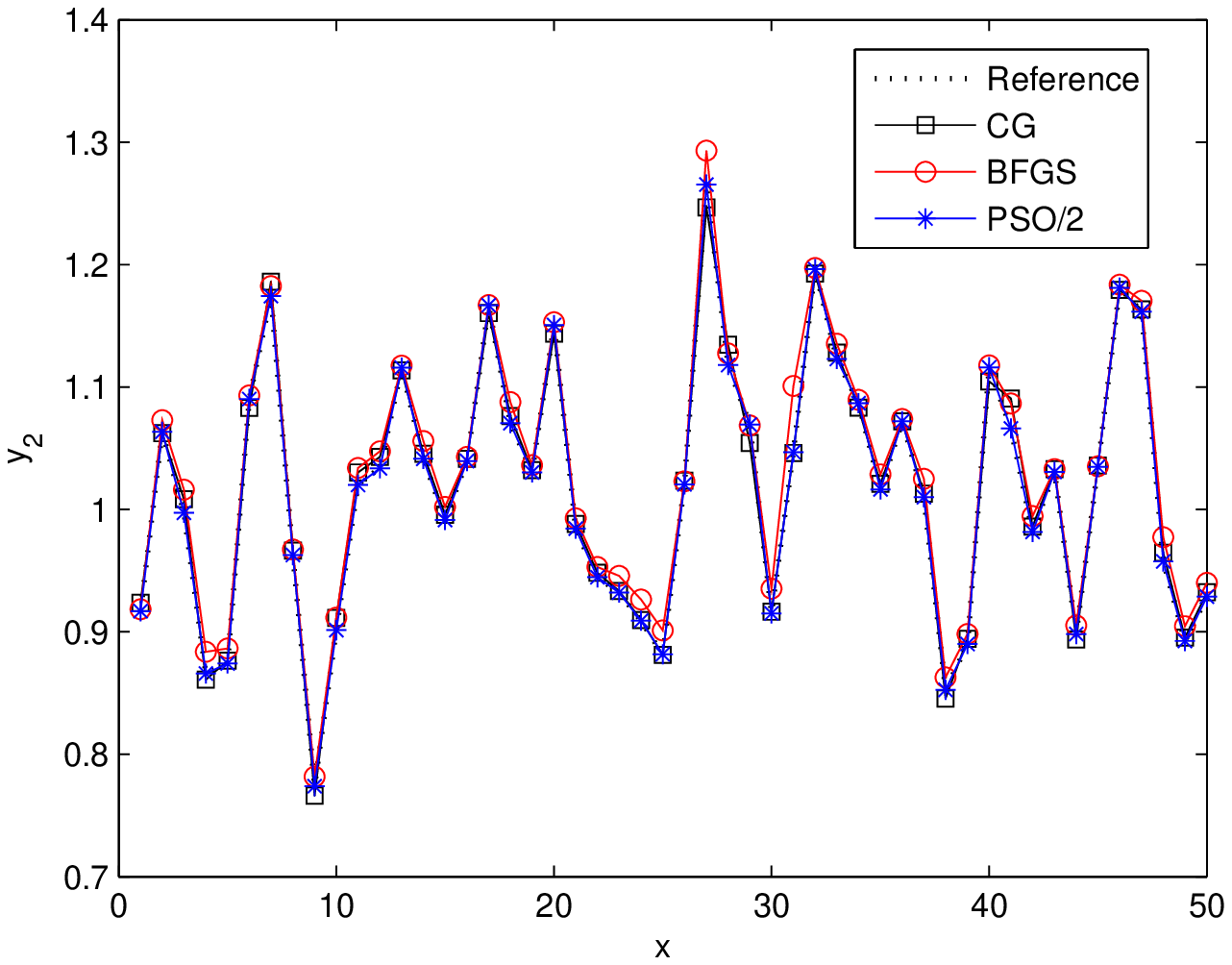}}
	\caption{Desired outputs and predicted outputs of \ac{cgp} models learned by PSO/2, CG and BFGS for the two-output modelling problem, where ``Linear" denotes $\mathbf{y}_2 = -\mathbf{y}_1$ and ``Nonlinear" represents $\mathbf{y}_2 = \textbf{exp}(\mathbf{y}_1)$.}
	\label{fig:mimoCGPlearn}
\end{figure*}

\textcolor{\colourred}{Table~\ref{tb:mimoCGPlearnResults-linear} and~\ref{tb:mimoCGPlearnResults-nonlinear} show the predicted output \ac{mse} of the \ac{cgp} models learned by the three different methods.
	These results show that PSO/2 outperforms the other two methods.
	This is confirmed by Figure~\ref{fig:mimoCGPlearn} which shows that
	the predicted outputs for PSO/2 are closer to the real outputs than for \ac{cg} and~\ac{bfgs}.}

\begin{figure*}[!t]
	\centering
	\subfigure[Case $1$]{
		\label{fig:certain}
		\centering
		\includegraphics[width=\multifigWidth\textwidth]{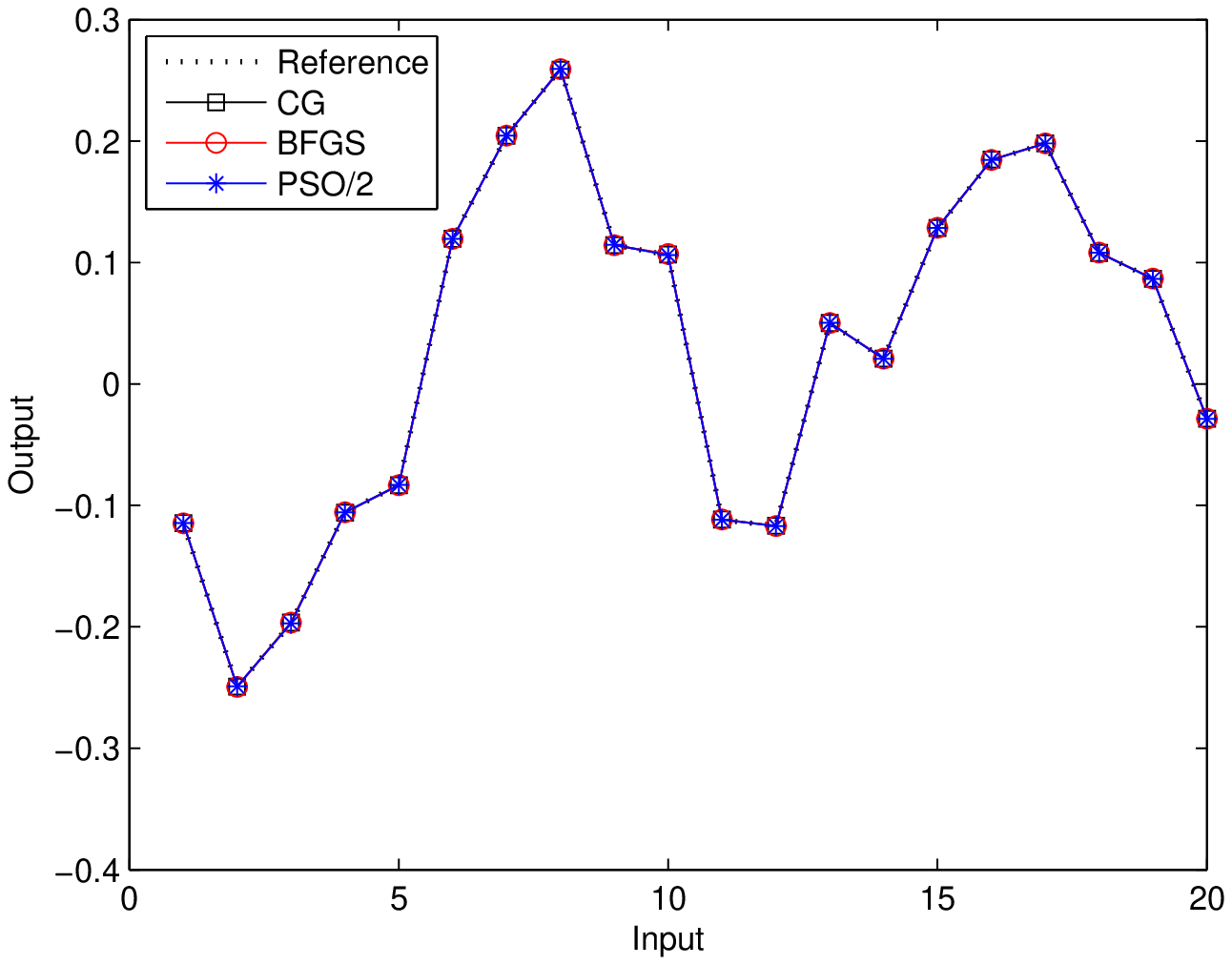}}
	\subfigure[Case $2$]{
		\label{fig:uncertain}
		\centering
		\includegraphics[width=\multifigWidth\textwidth]{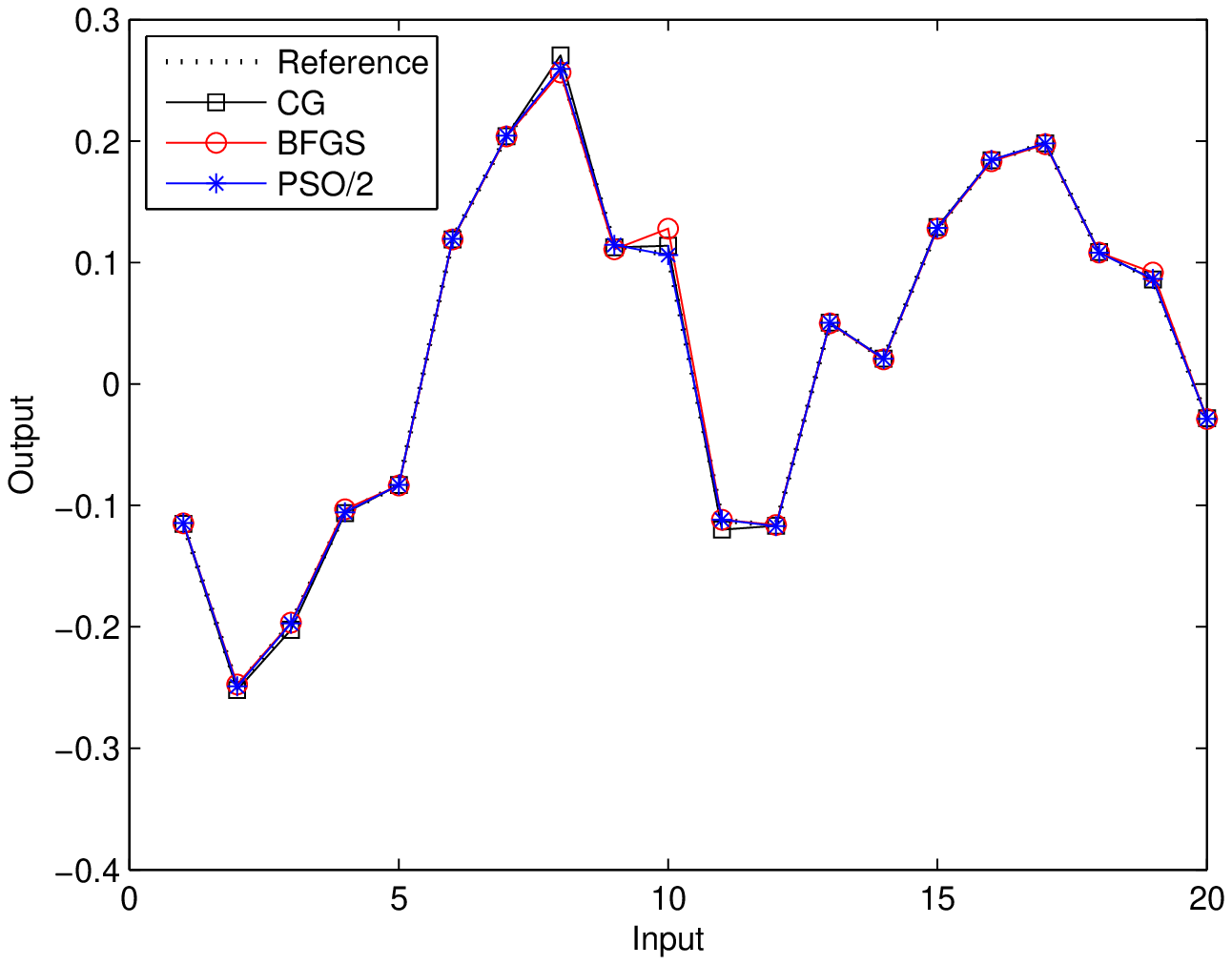}}
	\caption{Fitting curves between desired outputs and predicted outputs of~\ac{cgp} models learned by the proposed standard~\ac{pso} with~\ac{mse} fitness (denoted by PSO/2), CG and BFGS approaches for the both two cases}
	\label{fig:CurveFitting}
\end{figure*}

\subsection{Effects of Search Space}
\label{subsec:searchspace}
Next, we aim to determine the influence of using different search spaces in the problem of hyperparameter learning.
Two different cases are considered here.
The same single-output system as in Section~\ref{subsubsec:singleoutput} is used here.
In the first case (``case $1$"), it is assumed that prior knowledge of value ranges for the parameters in (\ref{eqn:particle}) is available.
More specifically,
\begin{equation}
\label{eqn:searchrange}
\begin{aligned}
\alpha_i, \beta_j, \nu_{d,i}, \upsilon_{q} \in \left[ 0,1 \right]
\end{aligned}
\end{equation}
where $\alpha_i$ and $\beta_j$ are the elements of the diagonal precision matrices $\mathbf{P_d}$ and $\mathbf{P_q}$ respectively.
\textcolor{\colourred}{In the second case (``case $2$"), a range of $[0, 100]$ that is much wider than (\ref{eqn:searchrange}) is used 
	for these parameters to indicate that we do not have any prior knowledge.}	

\textcolor{\colourred}{Prediction accuracies of the three methods are shown in Table~\ref{tb:sisoCGPlearnMSE}.
	They show that all three methods perform equally well with a well-defined search range.
	This is confirmed by Figure~\ref{fig:certain} for ``case $1$" where
	the predicted outputs of the three models are very close to desired one.
	But PSO/2 outperforms \ac{cg} and \ac{bfgs} when the search range is not well defined.
	Figure~\ref{fig:uncertain} shows that the models learnt by using \ac{cg} and \ac{bfgs}
	could not produce predicted outputs that follow the desired output as closely as the one learnt by PSO/2.}

\begin{table}[!t]
	\caption{The~\ac{mse} values of predicted outputs through using the~\ac{cgp} models learned by the proposed standard~\ac{pso} with~\ac{mse} fitness (denoted by PSO/2), \ac{cg} and~\ac{bfgs} in the single-output modelling problem}
	\label{tb:sisoCGPlearnMSE}
	\centering
	\begin{tabular}{c| c| c| c}
		\hline
		& PSO/2 & CG & BFGS \\
		\hline
		case $1$ & 3.9951e-08 & 1.4360e-07 & 1.3666e-07\\
		\hline
		case $2$ & 3.9951e-08 & 1.4135e-05 & 2.7007e-05\\
		\hline
	\end{tabular}
\end{table}

\begin{table*}[!t]
	\centering
	\caption{\ac{mse} of predicted outputs of \ac{cgp} models learned by the enhanced and standard \ac{pso} algorithms with the~\ac{nll} fitness, \ac{cg} and~\ac{bfgs} for LTV system modelling.}
	\begin{tabular}{c| c| c| c| c| c| c}
		\hline
		\multirow{2}{*}{} &\multicolumn{4}{c|}{PSO} & \multirow{2}{*}{CG} & \multirow{2}{*}{BFGS}\\ 
		\cline{2-5}
		& Standard & Gradient-based & Multi-Start& Hybrid & &\\
		\hline
		%Population & 20 & 20 & $\geq$20 & $\geq$20  & 20 $\times$ 500\\
		%\hline
		$y_1$ & 5.9673& 3.3801& 3.8991& 0.9717 & 9.0515& 10.8738\\
		\hline
		$y_2$ & 6.6911& 2.9001& 4.2333& 1.1231 & 8.6434& 9.8989\\
		\hline
	\end{tabular}
	\label{tb:mimoltvCGPlearn-ll}
\end{table*}
\begin{table*}[!t]
	\centering
	\caption{\ac{mse} of predicted outputs of \ac{cgp} models learned by the enhanced and standard \ac{pso} algorithms with \ac{mse} fitness, \ac{cg} and~\ac{bfgs} for LTV system modelling.}
	\begin{tabular}{c| c| c| c| c| c| c}
		\hline
		\multirow{2}{*}{} &\multicolumn{4}{c|}{PSO} & \multirow{2}{*}{CG} & \multirow{2}{*}{BFGS}\\ 
		\cline{2-5}
		& Standard & Gradient-based & Multi-Start& Hybrid & &\\
		\hline
		%Population & 20 & 20 & $\geq$20 & $\geq$20  & 20 $\times$ 500\\
		%\hline
		$y_1$ & 4.6271& 1.7861& 2.0847& 0.2703 & 10.9735& 9.8711\\
		\hline
		$y_2$ & 3.7600& 2.9174& 3.2472& 0.5074 & 9.0660& 9.9366\\
		\hline
	\end{tabular}
	\label{tb:mimoltvCGPlearn-mse}
\end{table*}

\subsection{Enhanced PSO Algorithms}
\textcolor{\colourred}{In this section, we evaluate the optimization performance of the three enhanced~\ac{pso} algorithms
	presented in Section~\ref{sec:psolearning}.
	The modelling of two non-trivial \ac{mimo} systems is considered.
	The results will be compared with those obtained by standard~\ac{pso}, \ac{cg} and~\ac{bfgs} algorithms.}

%============
%  LTV System
%============
\subsubsection{LTV System Modelling}
\label{subsubsec:LTV}

Consider a $2$-input-$2$-output~\ac{ltv} system~\parencite{PHD-Majji-TimeVaryingSystemIdentification-2009}
defined by,
\begin{equation}\label{eqn:ltvSystem} 
\begin{aligned}
\mathbf{\dot{x}}(t) & = \mathbf{A}(t)\cdot \mathbf{x}(t)+\mathbf{B}(t)\cdot \mathbf{u}(t)\\
\mathbf{y}(t) & = \mathbf{C}(t)\cdot \mathbf{x}(t)+\mathbf{D}(t)\cdot \mathbf{u}(t)
\end{aligned}
\end{equation}
where $\mathbf{A,B,C}$ and $\mathbf{D}$ are defined as:
\begin{equation}
\begin{aligned}
\mathbf{A}(t) &= 
\left[
\begin{array}{ccc}
0.3-0.9\Gamma_{1t} & 0.1 & 0.7\Gamma_{2t} \\
0.6\Gamma_{1t} & 0.3-0.8\Gamma_{2t} & 0.01 \\
0.5 & 0.15 & 0.6-0.9\Gamma_{1t} 
\end{array}
\right]       \\
\mathbf{B} &= 
\left[
\begin{array}{cc}
1 & 0 \\ 1 & -1 \\ 0 & 1
\end{array}
\right]
\mathbf{C} = 
\left[
\begin{array}{ccc}
1 & 0 & 1 \\ 1 & -1 & 0
\end{array}
\right]
\mathbf{D} = 0.1
\left[
\begin{array}{cc}
1 & 0 \\ 0 & 1
\end{array}
\right]
\end{aligned} 
\end{equation}
Matrix $\mathbf{A}$ has time-varying parameters $\Gamma_{1t} =\sin(10t)$ and $\Gamma_{2t} = \cos(10t)$.
The two control inputs are given by $u_1(t) = 0.5\sin(12t)$ and $u_2(t) = \cos(7t)$. They have zero initial conditions.

\textcolor{\colourred}{Using a sampling interval of $0.05s$,
	$200$ data records which include the inputs, states and outputs are generated.
	$60$ randomly selected samples are used for training, 
	and all $200$ samples are used for testing.
	The search range is $[0, 100]$ and \ac{cg} and~\ac{bfgs} algorithms are restarted $2000$ times.
	In addition, both the~\ac{nll} and~\ac{mse} are used as the fitness function for the enhanced and standard~\ac{pso} algorithms.}

\textcolor{\colourred}{The results are shown in Tables~\ref{tb:mimoltvCGPlearn-ll} and \ref{tb:mimoltvCGPlearn-mse}.
	In all cases,
	the $3$ enhanced~\ac{pso} methods perform better than the standard~\ac{pso}, \ac{cg} and \ac{bfgs}.
	In particular, the proposed hybrid~\ac{pso} produced the lowest \ac{mse}.
	In addition, comparing the corresponding entries in Tables~\ref{tb:mimoltvCGPlearn-ll} and \ref{tb:mimoltvCGPlearn-mse} suggests that
	using the output \ac{mse} as the fitness function for \ac{pso} algorithms seems to produce more accurate models.}  

\begin{figure*}[!t]
	\centering
	\subfigure[Log-Likelihood Fitness]{
		\label{fig:LTV_LL_Original}
		\centering
		\includegraphics[width=\multifigWidth\textwidth]{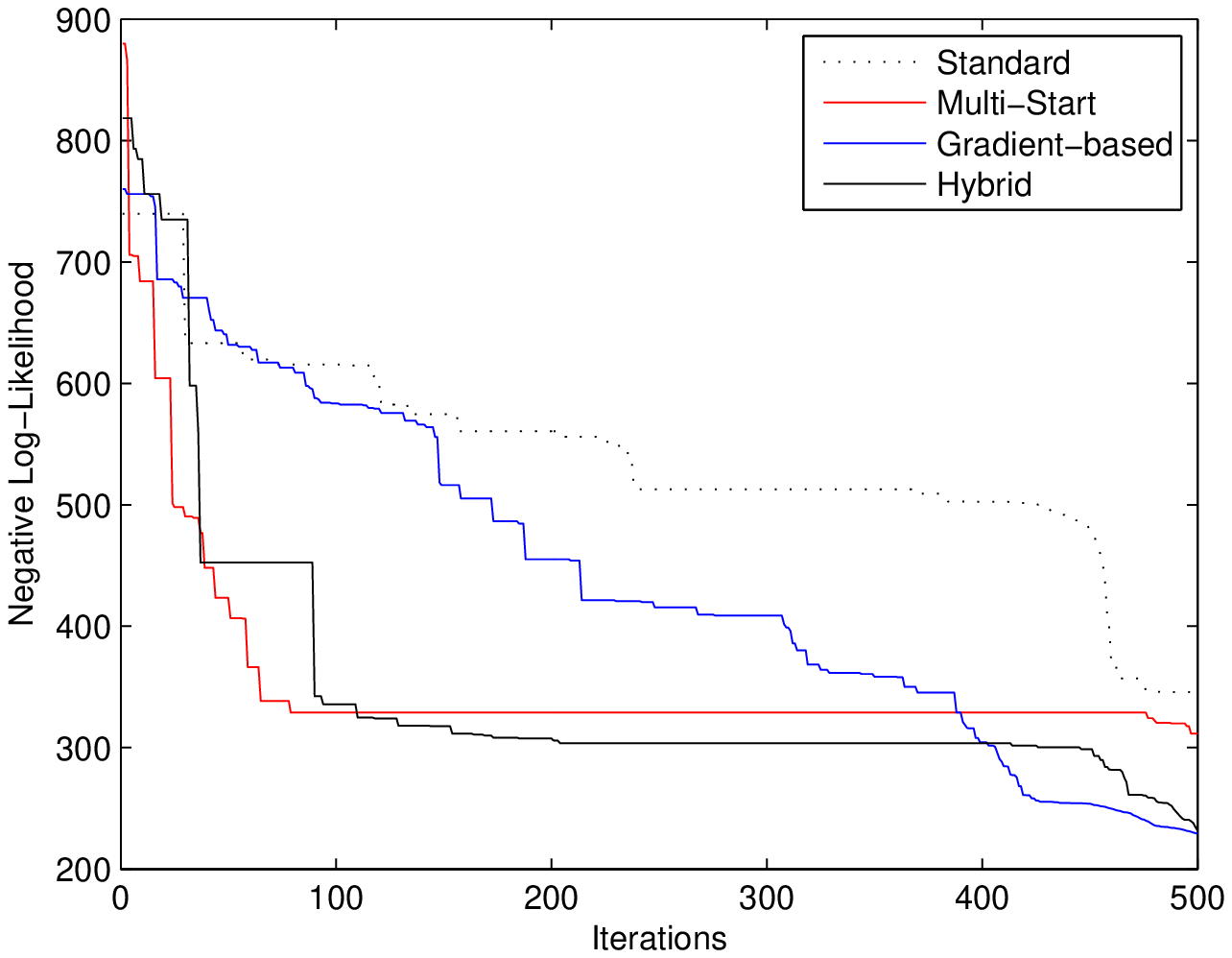}}
	\subfigure[MSE Fitness]{
		\label{fig:LTV_MSE}
		\centering
		\includegraphics[width=\multifigWidth\textwidth]{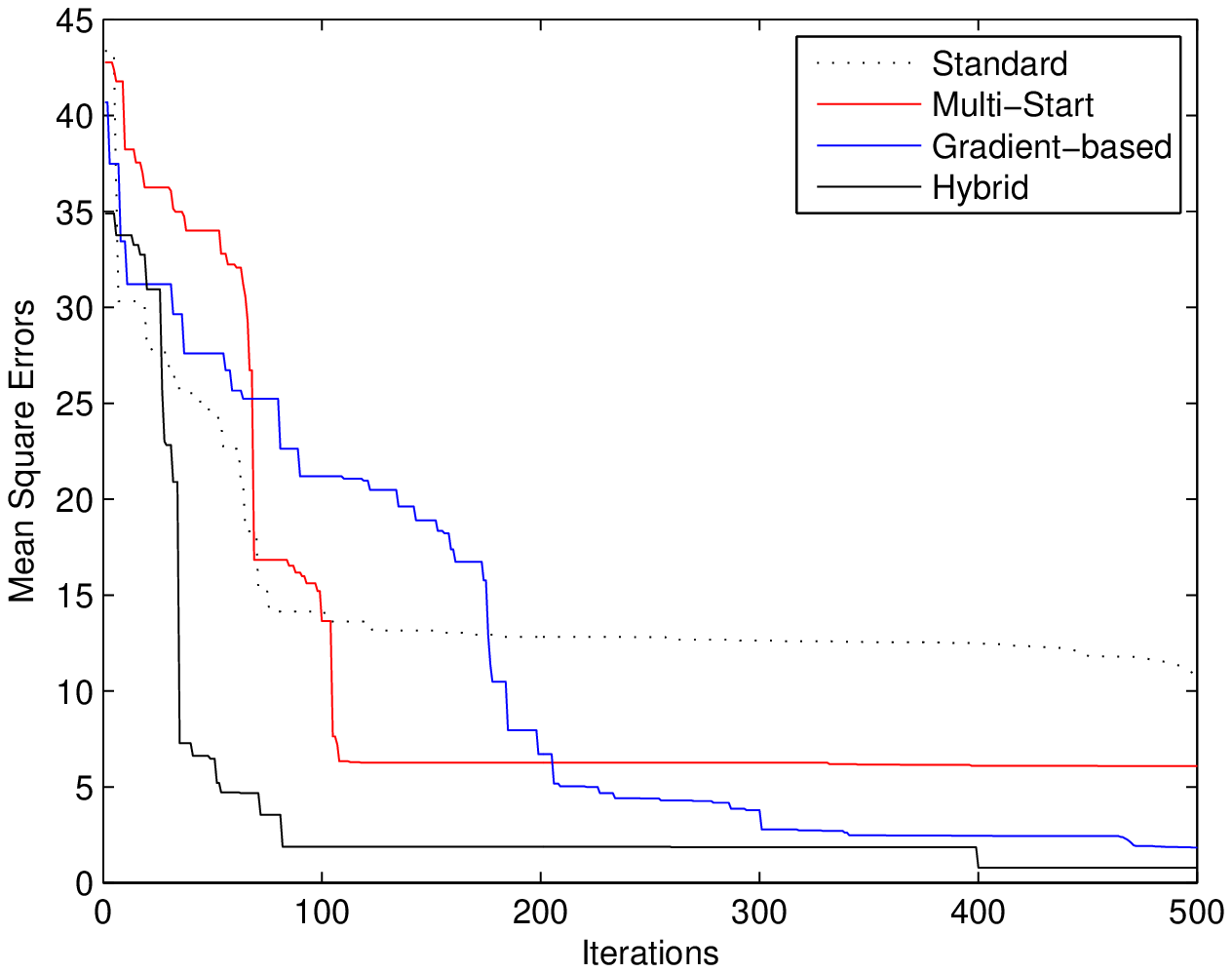}}
	\caption{Convergence behaviour of the proposed enhanced~\ac{pso} algorithms (multi-start, gradient-based and hybrid) and standard~\ac{pso} with the both~\ac{nll} and~\ac{mse} fitnesses in the modelling problem of the LTV system}
\end{figure*}

\textcolor{\colourred}{Figures~\ref{fig:LTV_LL_Original} and~\ref{fig:LTV_MSE} depict the convergence behaviours of the \ac{pso} algorithms.
	They show that
	the hybrid and multi-start~\ac{pso} algorithms perform a better search at the early stages 
	(approximately before $150$ iterations) than the standard and gradient-based~\acap{pso}.
	In addition,
	the hybrid and gradient-based~\ac{pso} methods are able to reach more optimal solutions than the multi-start and standard alternatives.
	It can therefore be concluded that the hybrid and gradient-based methods have better local search abilities (approximately after $400$ iterations) 
	than the other two approaches.
	Among the methods considered, the proposed hybrid~\ac{pso} method showed good local and global optimization performance.}

%============
%  NLTV System
%============
\subsubsection{NLTV System Modelling}

\begin{figure*}[!t]
	\centering
	\subfigure[``Step" Trajectory--Outputs $y(k)$]{
		\label{fig:Step_Y_PFDL}
		\centering
		\includegraphics[width=\multifigWidth\textwidth]{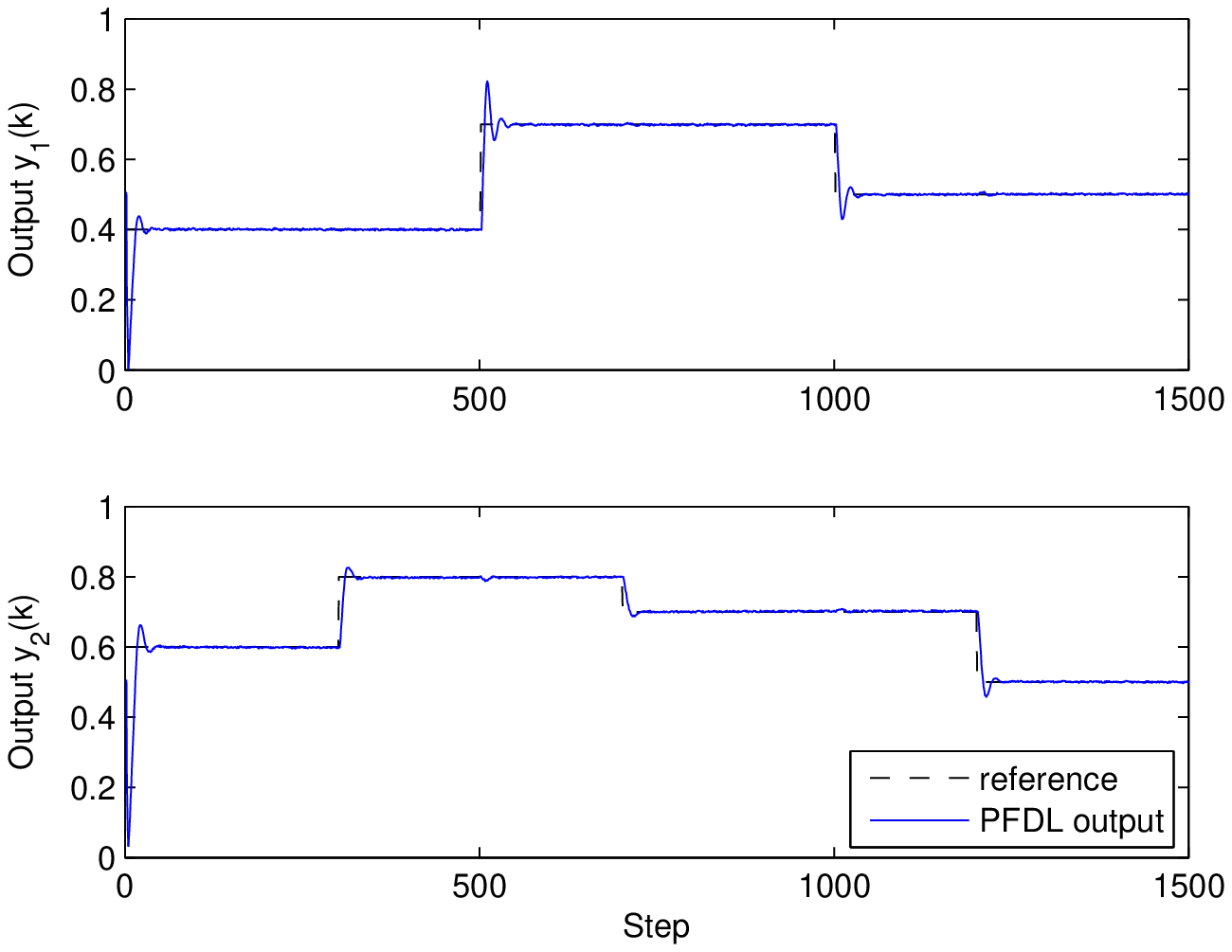}}
	\subfigure[``Step" Trajectory--Inputs $u(k)$]{
		\label{fig:Step_U_PFDL}
		\centering
		\includegraphics[width=\multifigWidth\textwidth]{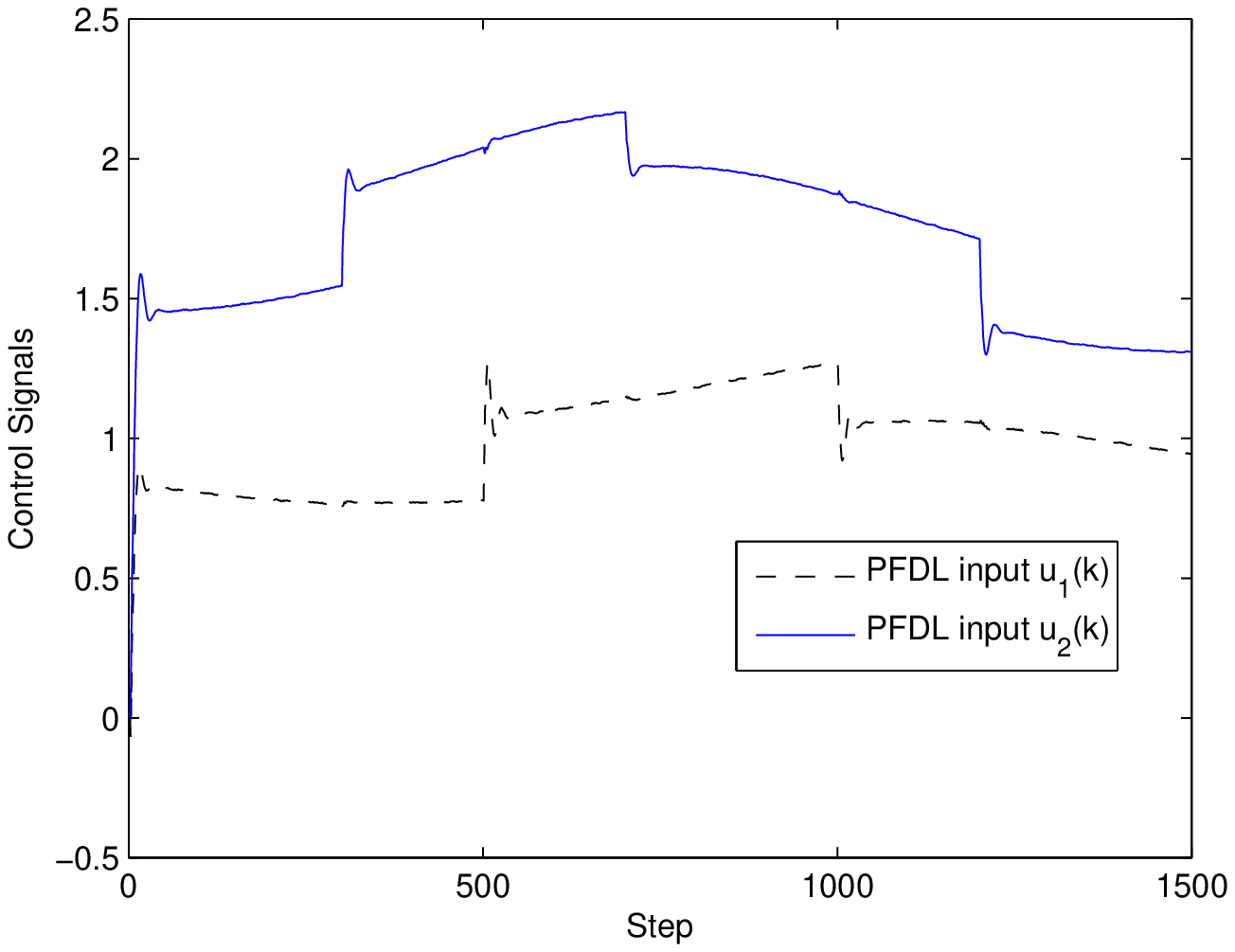}}\\
	\subfigure[``Curve" Trajectory--Outputs $y(k)$]{
		\label{fig:Curve_Y_PFDL}
		\centering
		\includegraphics[width=\multifigWidth\textwidth]{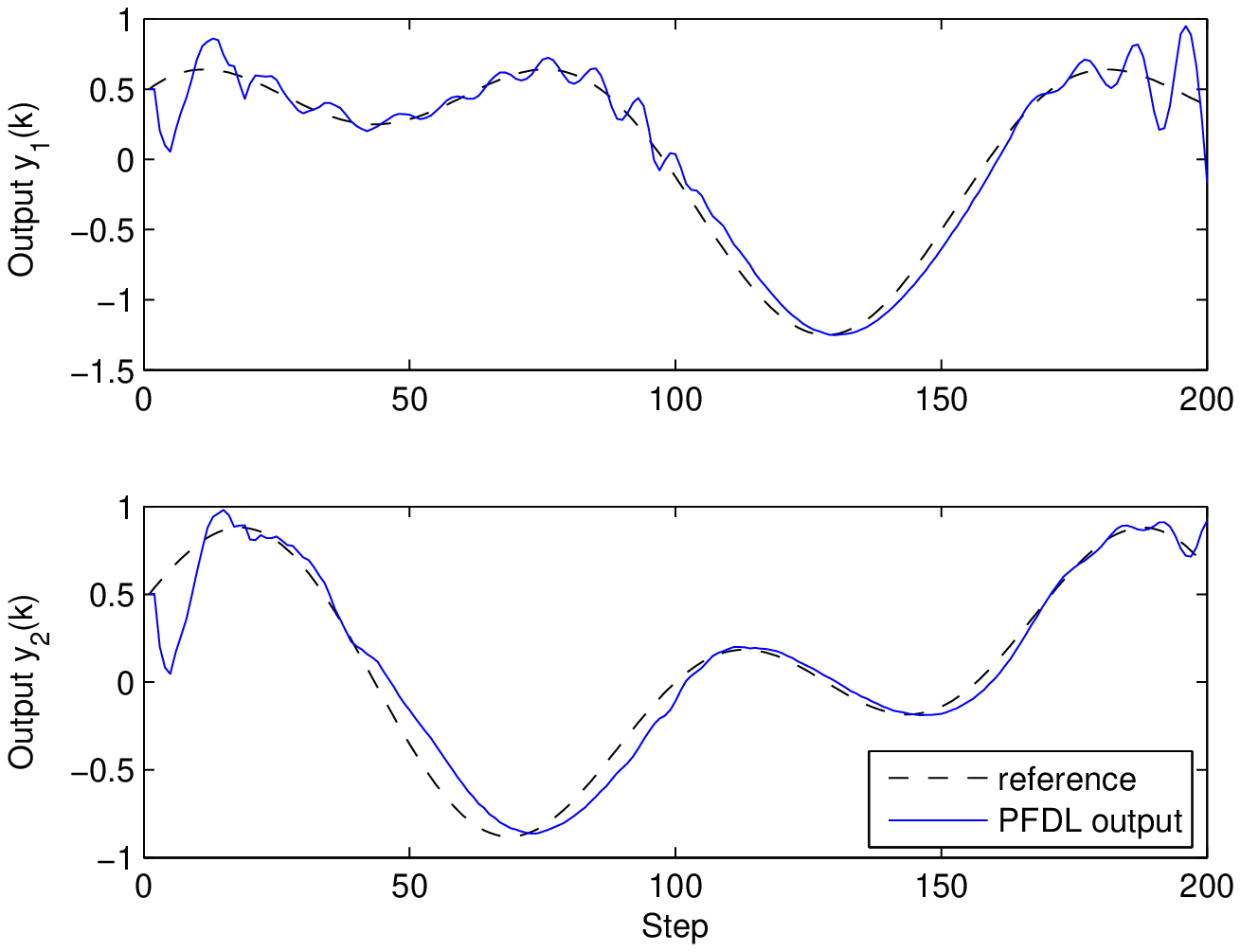}}
	\subfigure[``Curve" Trajectory--Inputs $u(k)$]{
		\label{fig:Curve_U_PFDL}
		\centering
		\includegraphics[width=\multifigWidth\textwidth]{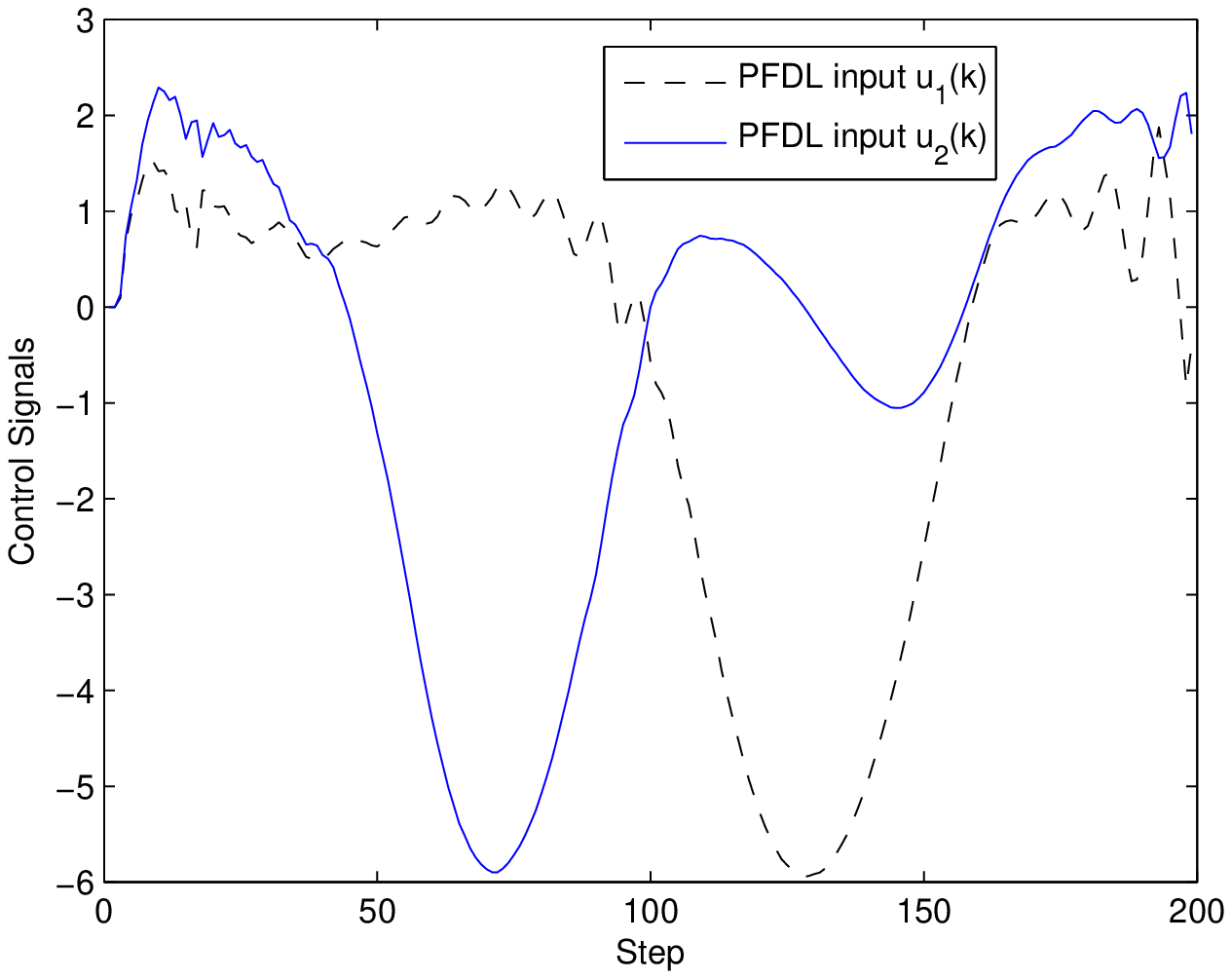}}
	\caption{Control inputs and outputs of using the~\ac{pfdl} appraoch for the two trajectories}
	\label{fig:reference_input_output}
\end{figure*}

The simulation in this section involves the~\ac{cgp} modelling of a~\ac{nltv} system controlled by a~\ac{pfdl} based~\ac{mfac} controller with the same parameters as in~\parencite{A-Hou-DataDrivenControl-MIMONLTV-2011}.
The $4$-input and $2$-output numerical system is described by,
\begin{equation}\label{eqn:nltvsysexample-2}
\begin{aligned}
x_{11}(k+1) =& \frac{x_{11}(k)^2}{1+x_{11}(k)^2} + 0.3x_{12}(k)\\
x_{12}(k+1) =& \frac{x_{11}(k)^2}{1+x_{12}(k)^2+x_{21}(k)^2+x_{22}(k)^2}+ a(k)u_1(k)\\
x_{21}(k+1) =& \frac{x_{21}(k)^2}{1+x_{21}(k)^2} + 0.2x_{22}(k)\\
x_{22}(k+1) =& \frac{x_{21}(k)^2}{1+x_{11}(k)^2+x_{12}(k)^2+x_{22}(k)^2}+ b(k)u_2(k)\\
y_1(k+1) =& x_{11}(k+1) + 0.005*\text{rand}(1)\\
y_2(k+1) =& x_{21}(k+1) + 0.005*\text{rand}(1)
\end{aligned}
\end{equation}
where the time-varying parameters are given by,
\begin{eqnarray}
\begin{aligned}
a(k) &= 1+0.1\sin(2\pi k/1500)\\
b(k) &= 1+0.1\cos(2\pi k/1500)
\end{aligned}
\label{eqn:sec2_nltvsim_tvparas}
\end{eqnarray}
This system is to track two trajectories.
One involves a ``Step" trajectory given by,
\begin{eqnarray}\label{eqn:StepTrack}
\begin{aligned}
y_1^*(k) =
\begin{cases}
0.4 & k \leq 500 \\
0.7 & 500 < k \leq 1000 \\
0.5 & 1000 < k \leq 1500
\end{cases}\\
y_2^*(k) =
\begin{cases}
0.6 & k \leq 300 \\
0.8 & 300 < k \leq 700 \\
0.7 & 700 < k \leq 1200 \\
0.5 & 1200 < k \leq 1500
\end{cases}
\end{aligned}
\end{eqnarray}
the other is ``Curve" trajectory specified by,
\begin{eqnarray}\label{eqn:CurveTrack}
\begin{aligned}
y_1^*(k) &= 0.75\sin(\frac{\pi k}{8})+0.5\cos(\frac{\pi k}{4})\\
y_2^*(k) &= 0.5\cos(\frac{\pi k}{8})+0.5\sin(\frac{\pi k}{4})
\end{aligned}
\end{eqnarray}
The same initial values of the system as \parencite{A-Zhang-OutputFeedbackControl-MIMO-2005} are used: $x_{11}(1)=x_{11}(2)=x_{21}(1)=x_{21}(2)=0.5$, $x_{12}(1)=x_{12}(2)=x_{22}(1)=x_{22}(2)=0$,
and $u_1(1)=u_1(2)=u_2(1)=u_2(2)=0$.
\textcolor{\colourred}{$1500$ and $200$ records are collected for the ``Curve" and ``Step" trajectories, respectively.
	In these simulations, we use a search range of $[0, 1]$ such that the optima or near-optima can be founded easier and faster than using $[0, 100]$.
	In addition, \ac{cg} and~\ac{bfgs} are again restarted $2000$ times}.

\textcolor{\colourred}{First,
	$40$ records are used for training the~\ac{cgp} models for both trajectories.
	The simulation results of using~\ac{mse} and~\ac{ll} in the~\ac{cgp} learning problem are given in Tables~\ref{tb:statisticNLTV-mse} and \ref{tb:statisticNLTV-ll}.
	Similar to the results obtained in Section~\ref{subsubsec:LTV}, the hybrid~\ac{pso} produces the lowest \ac{mse} values.
	In terms of the convergence behaviour, as shown in Figure~\ref{fig:mimoCGPlearn_NLTV},
	the hybrid algorithm convergences as fast as the multi-start~\ac{pso} at the early stage.
	At the same time, it is able to arrive at the most optimum values at the later stage.}

\begin{figure*}
	\centering
	\subfigure[``Step" Trajectory--NLL]{
		\label{fig:mimoCGPlearn_NLTV_Step_pso_LL}
		\centering
		\includegraphics[width=\multifigWidth\textwidth]{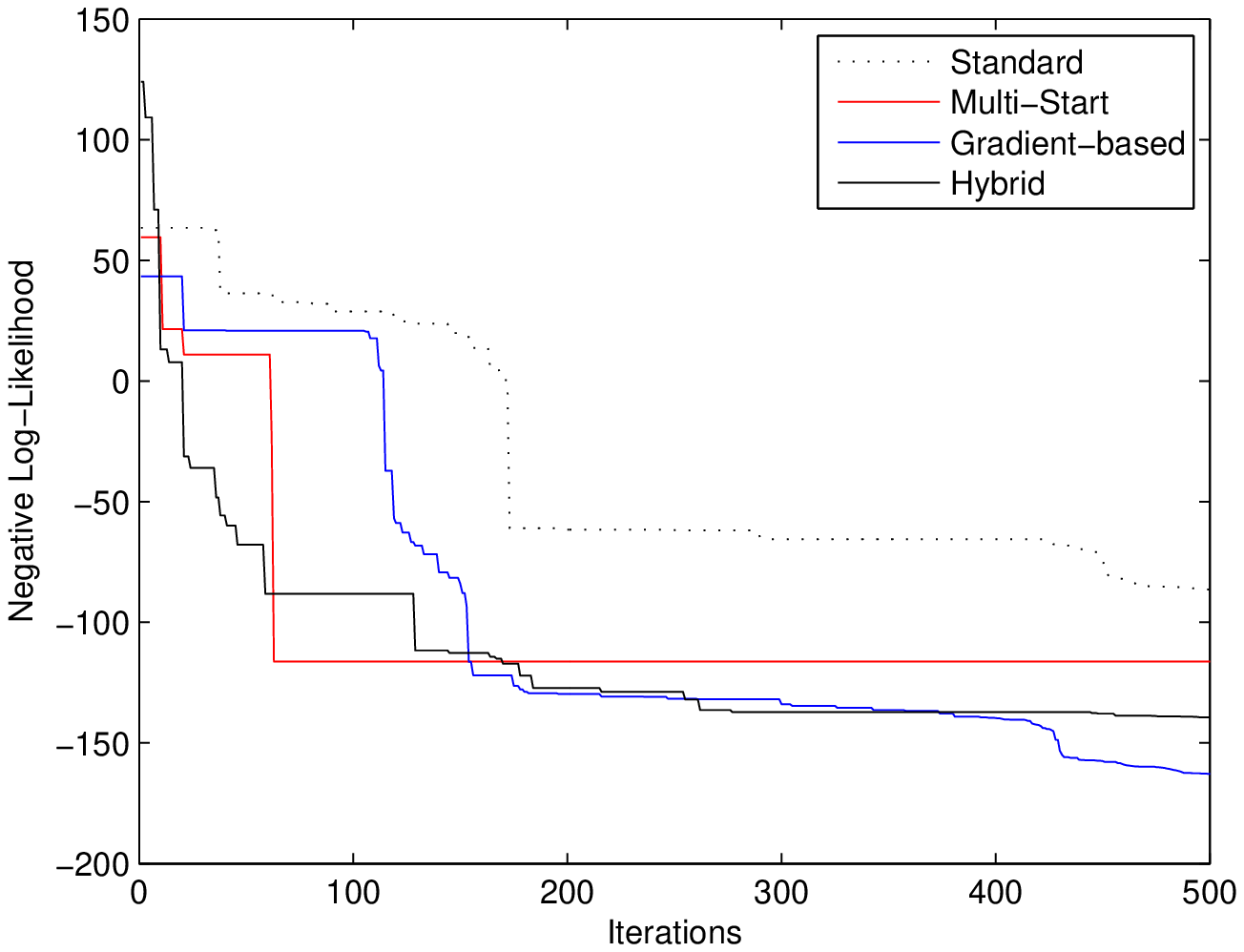}}
	\subfigure[``Curve" Trajectory--NLL]{
		\label{fig:mimoCGPlearn_NLTV_Curve_pso_LL}
		\centering
		\includegraphics[width=\multifigWidth\textwidth]{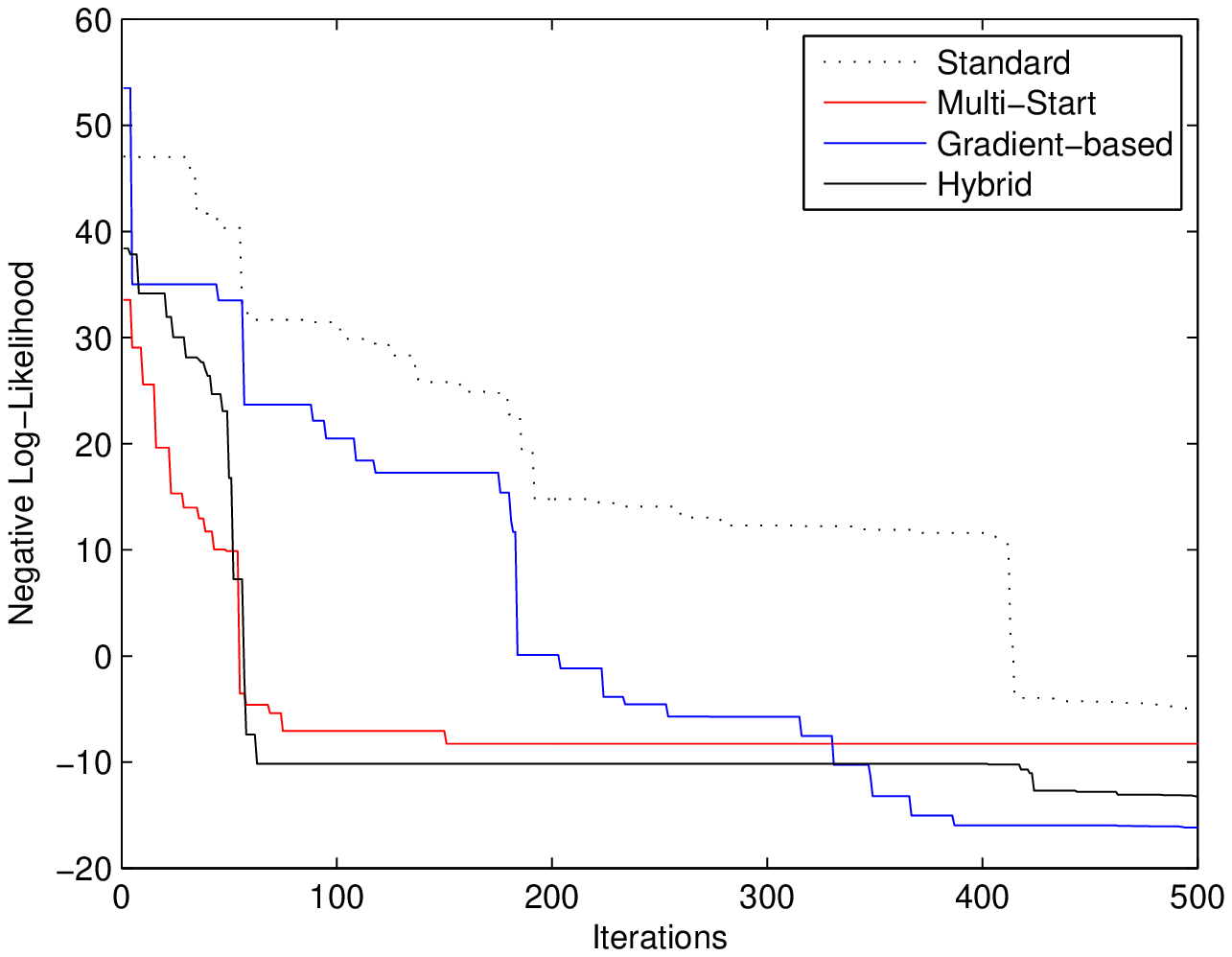}}
	\caption{Convergence behaviour of the proposed enhanced~\ac{pso} algorithms (multi-start, gradient-based and hybrid) and standard~\ac{pso} with the~\ac{nll} fitness in the modelling problem of the NLTV system}
	\label{fig:mimoCGPlearn_NLTV_NLL}
\end{figure*}
\begin{figure*}
	\centering
	\subfigure[``Step" Trajectory]{
		\label{fig:NLTV_MSE_Step}
		\centering
		\includegraphics[width=\multifigWidth\textwidth]{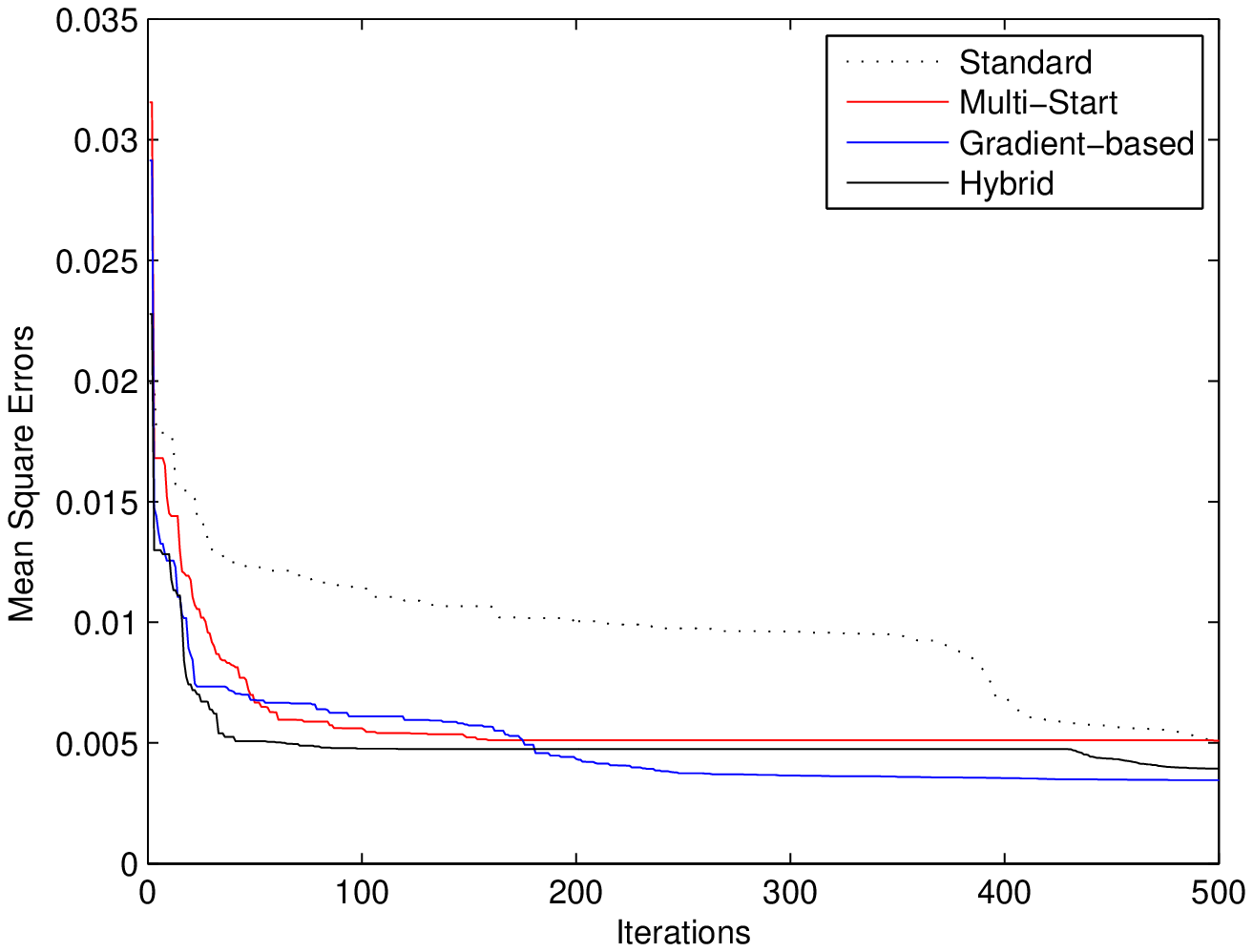}}
	\subfigure[``Curve" Trajectory]{
		\label{fig:NLTV_MSE_Curve}
		\centering
		\includegraphics[width=\multifigWidth\textwidth]{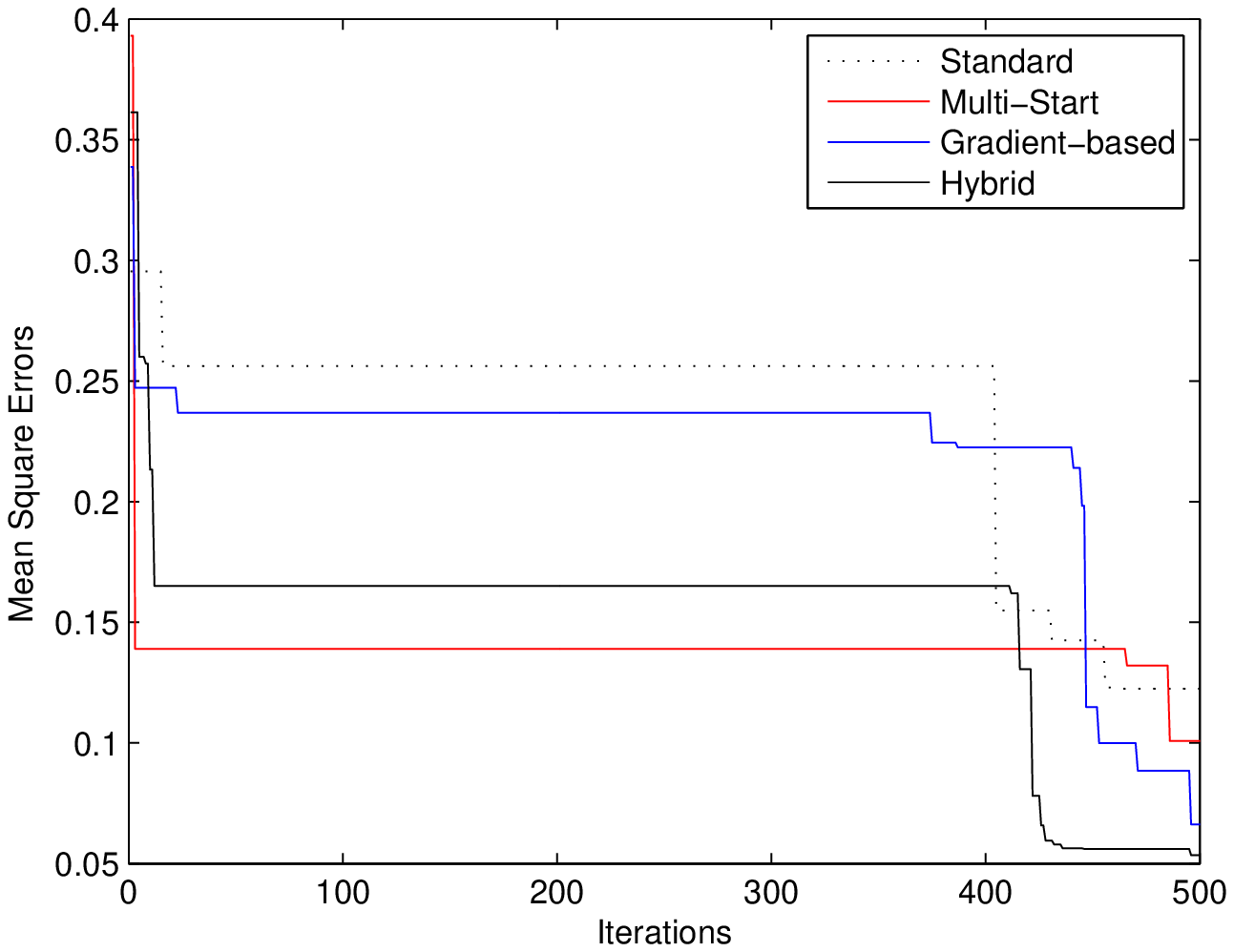}}
	\caption{Convergence behaviour of the proposed enhanced~\ac{pso} algorithms (multi-start, gradient-based and hybrid) and standard~\ac{pso} with the~\ac{mse} fitness in the modelling problem of the NLTV system}
	\label{fig:mimoCGPlearn_NLTV_MSE}
\end{figure*}

\begin{table*}
	\centering
	\caption{\ac{mse} of predicted outputs of the~\ac{cgp} models learned by the enhanced and standard \ac{pso} algorithms with \ac{mse} fitness, \ac{cg} and~\ac{bfgs} for  modellingthe NLTV system.}
	\begin{tabular}{c| c| c| c| c| c| c}
		\hline
		\multirow{2}{*}{} &\multicolumn{4}{c|}{PSO} &\multirow{2}{*}{CG} & \multirow{2}{*}{BFGS}\\ 
		\cline{2-5}
		& Standard & Gradient-based & Multi-Start& Hybrid & &\\
		\hline
		\multicolumn{7}{c}{``Step" Trajectory}\\
		\hline
		$y_1$ & 0.0837&0.0084 &0.0179 & 6.1475e-04& 0.1221 &0.5896 \\
		\hline
		$y_2$ & 0.0218&0.0062 &0.0337 & 7.6111e-04& 0.1273 &0.7785 \\
		\hline
		\multicolumn{7}{c}{``Curve" Trajectory}\\
		\hline
		$y_1$ & 0.3083&0.0417 &0.1594 & 0.0031& 0.1541 & 0.9657\\
		\hline
		$y_2$ & 0.1627&0.0402 &0.1098 & 0.0032& 0.2333 & 0.8811\\
		\hline
	\end{tabular}
	\label{tb:statisticNLTV-mse}
\end{table*}

\begin{table*}
	\centering
	\caption{\ac{mse} of predicted outputs of the~\ac{cgp} models learned by the enhanced and standard \ac{pso} algorithms with \ac{nll} fitness, \ac{cg} and~\ac{bfgs} for  modellingthe NLTV system.}
	\begin{tabular}{c| c| c| c| c| c| c}
		\hline
		\multirow{2}{*}{} &\multicolumn{4}{c|}{PSO} &\multirow{2}{*}{CG} & \multirow{2}{*}{BFGS}\\ 
		\cline{2-5}
		& Standard & Gradient-based & Multi-Start& Hybrid & &\\
		\hline
		\multicolumn{7}{c}{``Step" Trajectory}\\
		\hline
		$y_1$ & 0.0131&9.7544e-04 &0.0098 & 5.6981e-04& 0.8763 &1.2001 \\
		\hline
		$y_2$ & 0.0087&9.5770e-04 &0.0012 & 2.1458e-04& 0.8001 &0.9899 \\
		\hline
		\multicolumn{7}{c}{``Curve" Trajectory}\\
		\hline
		$y_1$ & 0.4681&0.1257 &0.3877 & 0.0977& 0.3048 & 0.3008\\
		\hline
		$y_2$ & 0.5002&0.1366 &0.4102 & 0.0854& 0.1130 & 0.1339\\
		\hline
	\end{tabular}
	\label{tb:statisticNLTV-ll}
\end{table*}

\begin{table*}[!t]
	\centering
	\caption{The comparison of learning the NLTV system through using the proposed~\ac{mse} fitness hybrid~\ac{pso} with different training data sizes in terms of the computational time and the~\ac{mse} values of predicted outputs of obtained~\ac{cgp} models}
	\setlength{\tabcolsep}{16pt}
	\begin{tabular}{c|c|c|c}
		\hline
		\multirow{1}{*}{Training} &\multicolumn{2}{c|}{MSE} &\multirow{2}{*}{Time(seconds)}\\
		\cline{2-3}
		Data Size & $y_1$ & $y_2$ & \\
		\hline
		\multicolumn{4}{c}{``Step" Trajectory}\\
		\hline
		20 & 0.0377 & 0.0511 & $\approx$12s\\
		\hline
		40 & 6.1475e-04 & 7.6111e-04 & $\approx$17s\\
		\hline
		100 & 1.1292e-04 & 1.3543e-04 & $\approx$31s\\
		\hline
		200 & 1.3411e-05 & 1.8854e-05 & $\approx$110s\\
		\hline
		\multicolumn{4}{c}{``Curve" Trajectory}\\
		\hline
		25 & 0.0562 & 0.0665 & $\approx$14s \\
		\hline
		50 & 0.0031 & 0.0032 & $\approx$18s \\
		\hline
		75 & 0.0012 & 0.0011 & $\approx$23s \\
		\hline
		100 & 1.1712e-04 & 1.9201e-04 & $\approx$29s \\
		\hline
	\end{tabular}
	\label{tb:PerformNumbers}
\end{table*}

\textcolor{\colourred}{The effect of the training data size on model accuracy 
	for the hybrid~\ac{pso} algorithm with~\ac{mse} fitness function is now evaluated.
	Training data are chosen from the control intervals shown in Figures~\ref{fig:Step_U_PFDL} and \ref{fig:Curve_U_PFDL}.
	The results of using different training sizes are shown in Table~\ref{tb:PerformNumbers}.
	As expected, model accuracy improves as the training data size increases.
	However, the algorithm runtime increases exponentially with data size.
	Interestingly, for the ``Step" trajectory where the outputs are piecewise constant, the system can be modelled with
	far fewer training data compared with the ``Curve" trajectory with continuously smooth outputs.}

\section{Conclusion}
\label{sec:conclude}

The hyperparameters of the~\ac{gp} models are conventionally learnt by minimizing the~\ac{nll} function.
This typically leads to an unconstrained nonlinear non-convex optimization problem that is usually solved by using the~\ac{cg} algorithm.
Three enhanced~\ac{pso} algorithms have been proposed in this chapter to improve the hyperparameter learning for~\ac{cgp} models of~\ac{mimo} systems.
They make use of gradient-based technique and also combine it with the multi-start technique.
Using numerical~\ac{ltv} and~\ac{nltv} systems, we have shown that these algorithms are more effective in avoiding getting stuck in local optima.
Hence they are able to produce more accurate models of the systems.
Results showed that the hybrid~\ac{pso} algorithm allows the faster convergence and produces the more accurate models.
These algorithms also use the~\ac{mse} of model outputs rather than the \ac{ll} function as the fitness function of optimization problems.
This enables us to assess the quality of intermediate solutions more directly.

\printbibliography

\end{document}